\PassOptionsToPackage{table}{xcolor}
\documentclass[onecolumn]{robbyant}           %

\usepackage{makecell}
\usepackage{wrapfig}
\usepackage{subcaption}
\usepackage{booktabs,array,tabularx,colortbl}
\usepackage[table]{xcolor}
\usepackage{multirow}
\usepackage{longtable}
\usepackage{amsthm}

\definecolor{pie1}{HTML}{5B8FF9}
\definecolor{pie2}{HTML}{5AD8A6}
\definecolor{pie3}{HTML}{F6BD16}
\definecolor{pie4}{HTML}{E86452}
\definecolor{pie5}{HTML}{6DC8EC}
\definecolor{pie6}{HTML}{945FB9}
\definecolor{pie7}{HTML}{FF9845}
\definecolor{pie8}{HTML}{1E9493}
\definecolor{pie9}{HTML}{FF99C3}
\definecolor{TactileDefault}{HTML}{E6F2FC}

\metadata[Website]{\url{https://technology.robbyant.com/lingbot-touch}}
\metadata[Github]{\url{https://github.com/robbyant/lingbot-touch}}
\metadata[Checkpoints]{\url{https://huggingface.co/collections/robbyant/lingbot-touch}}
\newcommand{\methodname}{LingBot-Touch}
\newcommand{\method}{\texttt{\methodname}\xspace}
\newcommand{\lingbotvla}{\texttt{LingBot-VLA}\xspace}
\newcommand{\objname}{Tactile-JEPA}
\newcommand{\obj}{\texttt{\objname}\xspace}

\title{Learning Heterogeneous Tactile Representation for General Dexterous Manipulation}

\author{
\begin{center}
    Bing Han \quad
    Yichuan Peng \quad
    Haoyu Dong \quad
    Yuanmeng Hu \quad
    Yongyi Jia \quad
    Fangjing Wang \quad
    Kecheng Zheng \quad
    \\[2pt]
    Yiming Zhang \quad
    Yanhong Zeng \quad
    Xiaoqing Zhang \quad
    Chen Song \quad
    Rundong Zou \quad
    Anqi Dong \quad
    \\[2pt]
    Yuan Guan\quad
    Ka Leong Cheng \quad
    Xing Zhu\quad
    Yujun Shen\quad 
    Nan Xue \quad
    Wei Ning$^\dagger$\quad 
    \\[12pt]
    $^{\dagger}$Project Lead
\end{center}
}

\begin{document}

\abstract{
Tactile sensing is an indispensable modality for contact-rich dexterous manipulation, a capability that hints at the ultimate frontiers of embodied intelligence.
However, commercially available tactile sensors are still at a relatively early stage of development, with each type of sensor offering irreplaceable strengths alongside its own drawbacks.
A key challenge is thus to free tactile understanding from sensor-specific physics, so that embodied models can effectively leverage heterogeneous tactile data.
To this end, we develop \method, a tactile foundation model characterized by the following properties.
(i) Our model yields a \textbf{\textit{unified representation}} across different tactile sensors (\textit{i.e.}, piezoresistive, magnetic, and visuotactile sensors), and is easily extensible to new sensor types.
(ii) The resulting tactile tokens are \textbf{\textit{plug-and-play}}: they can be seamlessly integrated into policy models across paradigms, from diffusion policy to vision-language-action (VLA) models such as $\pi_{0.5}$ and \lingbotvla, consistently enhancing manipulation dexterity.
(iii) Extensive experiments, spanning simulation and real-robot deployments across multiple hardware platforms, demonstrate that our tactile representation substantially \textbf{\textit{benefits complex manipulation tasks}} while maintaining strong computational efficiency.
We believe the findings in this work demonstrate the potential of leveraging cross-sensor data to scale tactile learning, and offer encouraging insights for the design of future hardware and embodied models.
}

\maketitle
\justifying

\begin{figure*}[t]
    \centering
    \includegraphics[width=1.0\linewidth]{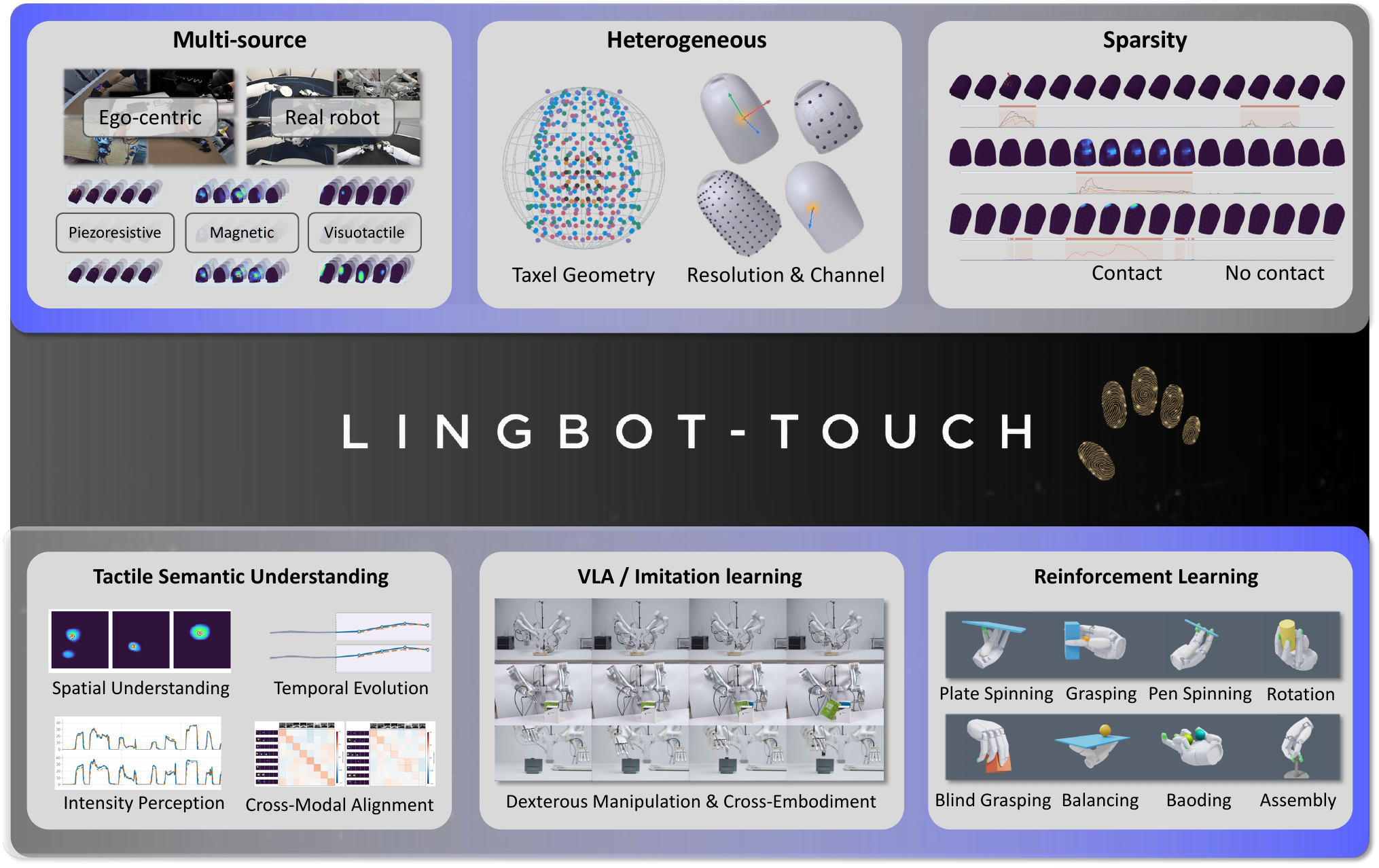}
    \caption{\textbf{Overview of \methodname.} \textit{Top:} what tactile data look like at scale. Our 5{,}000-hour corpus is multi-source, pooling egocentric recordings from human tactile gloves with teleoperated robot hands across piezoresistive, magnetic, and optical sensing (left); heterogeneous in taxel geometry, resolution, and channels, which we reconcile in a canonical fingertip space (middle); and sparse, with most taxels silent most of the time (right). \textit{Bottom:} what one encoder trained on this corpus enables. Frozen features recover contact force, location, extent, and dynamics and align with vision (left); the same tokens plug into imitation-learning and VLA policies on real dual-arm hands (middle) and into reinforcement learning in simulation (right). \methodname{} transfers to a sensor model unseen in pretraining without retraining, and its 5.46M-parameter Tiny variant adds under 0.2\% to the parameters of the policies it augments.}
    \label{fig1}
\end{figure*}

\section{Introduction}
\label{sec:intro}

Dexterous manipulation with multi-fingered hands is among the most remarkable abilities of humans, and giving robots the same ability is a central goal of physical intelligence. Much of that ability rests on touch: with fingertip sensation blocked, people struggle with tasks as simple as striking a match or buttoning a shirt, even with vision intact~\cite{johansson2009coding}. Robots, by contrast, still manipulate objects almost entirely by sight~\cite{chi2023diffusionpolicy,pmlr-v305-black25a, wu2026foundation}, and sight cannot tell them how hard they are pressing, whether a grasp is slipping, or what is happening under an occluding finger.

Why not touch? Not for lack of sensors. Piezoresistive, magnetic, and visuotactile sensors are all commercially available (Fig.~\ref{fig1}, top), but touch remains an early-stage modality that lacks the three ingredients that let vision and language scale. There is no common input format: each sensor type defines its own channels, layout, resolution, and sampling rate~\cite{zhao2025t3,bi2026htt}, so a reading from one sensor means nothing to a model trained on another. There is no large curated corpus, because data from one sensor cannot be pooled with data from the next and every hour requires a sensorized hand touching the world. And there is no established recipe: evidence that touch helps robot learning~\cite{higuera2025sparsh,zhao2025t3} is scattered across sensor-specific studies and individual tasks, with no shared representation to compare them on.

The three gaps share a root: tactile data are tied to the physics of the sensor that produced them. Contact itself is not. A fingertip pressing on an edge is the same event whether a magnetometer, a resistive grid, or a camera behind a gel records it, so a representation of contact can be shared once sensor-specific physics is removed from the input. In this paper, we present \method, a tactile foundation model that learns such a representation from the three dominant sensing principles, piezoresistive, magnetic, and optical, with one encoder, and extends to new sensors without retraining. Three design decisions make this possible. We map every sensor's taxels into a canonical fingertip space, so that a model sees positions on a finger rather than indices in a sensor-specific array (Sec.~\ref{sec:method:tjepa}). We train with a joint-embedding predictive architecture~\cite{assran2023ijepa,vjepa} that predicts representations of masked taxels in latent space, which suits signals that are sparse and carry no labels (Sec.~\ref{sec:method:tjepa}). And we assemble a 5{,}000-hour corpus from human tactile gloves and teleoperated robot hands, the first at a scale where representation learning for touch becomes possible (Sec.~\ref{sec:method:corpus}).

Trained once on this corpus, \method behaves the way a foundation model should (Fig.~\ref{fig1}, bottom). Its frozen features recover contact force, location, extent, and short-term dynamics from fingertip signals and retrieve the matching video frame from touch alone, on an optical hand seen in pretraining and on a magnetic hand whose sensor model was not; on the latter it outperforms FTP-1~\cite{ftp1} fitted with a sensor-specific adapter on all five probing tasks (Secs.~\ref{sec:ContactPhysics}, \ref{sec:exp:scalingtransfer}). Its tokens plug into existing policies without changing them: on 26 real-world dual-arm tasks, adding the frozen 5.46M-parameter encoder raises the success rate of Diffusion Policy~\cite{chi2023diffusionpolicy}, $\pi_{0.5}$~\cite{pmlr-v305-black25a}, \lingbotvla~\cite{wu2026foundation}, and T-Rex~\cite{ niu2026trextactilereactivedexterousmanipulation}
by 14 to 26 percentage points, while feeding the same policies raw tactile signals yields at most about half that gain~(Sec.~\ref{sec:exp:robot}). In simulation, reinforcement learning on \method features succeeds on 69\% of attempts across twelve dexterous tasks, against 46\% with features from a task-specific CNN and 21\% without touch, and with far lower variance across seeds (Sec.~\ref{sec:exp:robot:sim}). The representation also scales: the Small model improves monotonically from 1K to 5K hours of pretraining on every probing task and both hands (Sec.~\ref{sec:exp:scalingtransfer}), data from one sensor improves the representation of another (Sec.~\ref{sec:Multi-Source}), and latent prediction outperforms signal reconstruction by a wide margin, with a VAE objective even degrading as data grow (Sec.~\ref{sec:exp:vae}).

Our contributions are as follows.
\begin{itemize}[leftmargin=*]
    \item \textbf{A tactile corpus at scale.} 5{,}000 hours of recordings, 464K episodes, and 1.33 billion frames across piezoresistive, magnetic, and optical sensing, collected from human tactile gloves and teleoperated robot hands over 24 task categories.
    \item \textbf{A tactile foundation model.} \method combines a canonical fingertip space, modality-specific tokenizers, and a shared Haptic Transformer, trained with \obj, a latent-prediction objective adapted to sparse, unlabeled tactile signals through a mask warm-up and layout augmentation. One encoder serves all three sensing principles and transfers to an unseen sensor model without retraining.
    \item \textbf{Evidence across the stack.} Frozen-encoder probing of contact physics on two hands, 26 real-world manipulation tasks with four policy families, and twelve reinforcement-learning tasks in simulation, together with scaling, multi-source, and objective analyses that explain where the gains come from.
    \item \textbf{An open release.} The inference code, a small set of sample tactile recordings, and checkpoints for the Tiny and Small variants pretrained on approximately 5{,}000 hours of tactile data will be released. These resources will help researchers integrate the tactile encoder into downstream robot learning systems.
\end{itemize}

\section{Related Work}
\label{sec:related}

\subsection{Tactile Sensing}
\label{sec:related:sensing}

Tactile sensors capture contact geometry, forces, and interaction dynamics through diverse transduction mechanisms. Magnetic sensors, such as ReSkin~\cite{bhirangi2022reskin} and AnySkin~\cite{bhirangi2025anyskin}, sense contact through deformation-induced changes in magnetic fields; PaXini's commercial systems also employ Hall-effect sensing~\cite{paxiniPXCapIII}. Optical tactile sensors, including GelSight~\cite{yuan2017gelsight}, DIGIT~\cite{lambeta2020digit}, and the fingertip sensors integrated into Sharpa Wave~\cite{sharpa2025wave}, use internal imaging to capture contact-induced deformation. Resistive and capacitive sensors encode contact through changes in electrical resistance or capacitance, exemplified by the piezoresistive Scalable Tactile Glove (STAG)~\cite{sundaram2019stag} and the conformable capacitive electronic skin DexSkin~\cite{wistreich2025dexskin}.

These technologies observe shared properties of physical contact through different measurement mechanisms, producing signals that vary in channel semantics, spatial resolution, sampling rate, and sensing layout. Similar interactions can therefore yield substantially different observations across sensors. \method addresses this heterogeneity through a common interface that accommodates diverse measurements while preserving contact geometry and temporal structure, enabling unified spatiotemporal contact representations across sensing modalities.

\subsection{Tactile Representation Learning}
Self-supervised representation learning has driven recent advances in large language and multimodal models, contributing to the emergence and development of world models, vision-language-action (VLA) models, and world-action models. Relevant approaches can be broadly grouped into joint-embedding architectures, such as DINO~\cite{caron2021dino}, CLIP~\cite{radford2021clip}, SigLIP~\cite{zhai2023siglip}, and VideoCLIP~\cite{xu2021videoclip}; generative architectures, such as MAE~\cite{he2022mae} and VideoMAE~\cite{tong2022videomae}; and joint-embedding predictive architectures, including I-JEPA~\cite{assran2023ijepa}, V-JEPA~\cite{vjepa}, V-JEPA 2~\cite{vjepa2}, LeJEPA~\cite{balestriero2025lejepa}, and V-JEPA 2.1~\cite{vjepa21}.

Tactile representation learning adapts these approaches to tactile observations. Sparsh~\cite{higuera2025sparsh} investigates reconstruction, self-distillation, and joint-embedding prediction for optical tactile sensing. T3~\cite{zhao2025t3} combines sensor-specific encoders with a shared Transformer backbone to enable transfer across optical tactile sensors and tasks. UniTouch~\cite{yang2024unitouch} connects tactile representations to pretrained multimodal embeddings, while AnyTouch~\cite{feng2025anytouch} combines masked modeling, multimodal alignment, and cross-sensor matching to learn static and dynamic tactile features. AnyTouch 2~\cite{feng2026anytouch2} further emphasizes force perception and temporal dynamics across optical tactile sensors. Sparsh-skin~\cite{sharma2025sparshskin} learns contextual representations from magnetic tactile histories and sensor geometry, whereas Sparsh-X~\cite{higuera2025sparshx} integrates image, audio, motion, and pressure signals from Digit 360~\cite{lambeta2024digit360}. HTT~\cite{bi2026htt} addresses heterogeneous optical and array-based sensors through modality-specific reconstruction and representation alignment between paired sensors. FTP-1~\cite{ftp1} maps heterogeneous observations into morphology-aware tokens for foundation-policy pretraining and supports adaptation to new hardware through sensor-specific encoders.

To address the high sparsity, low signal-to-noise ratios, and multi-source heterogeneity of tactile observations, we introduce \method, a tactile foundation model that combines a multi-branch tokenizer with a shared Haptic Transformer to learn unified, reusable representations of contact physics across sensors and heterogeneous layouts.

\section{\methodname}
\label{sec:method}
\method learns actionable tactile representations for dexterous manipulation policies from data. Two challenges stand in the way. First, tactile data are scarce. Unlike vision and language, no tactile corpus exists today at the scale that representation learning requires. We address this directly by collecting 5{,}000 hours of tactile recordings (\cref{sec:method:corpus}). Because this effort spans human tactile gloves and robot hands with different sensors, the resulting data are heterogeneous in transduction principle, taxel layout, resolution, and sampling rate, which we handle through a canonical geometric interface and modality-specific tokenizers that feed a shared Haptic Transformer (\cref{sec:method:tjepa}). Second, these recordings come with at most coarse task-level descriptions, and nothing that labels the contact physics inside a clip, such as the forces exerted, where and how widely contact occurs, or when it begins and ends. The model must therefore discover the structure of contact on its own. We address this with a self-supervised objective that predicts contact in latent space, together with a mask warm-up that keeps training stable on sparse signals (\cref{sec:method:tjepa}), and with layout augmentation that lets the learned representation transfer across sensors (\cref{sec:method:aug}).

\paragraph{Problem formulation.} We represent a normalized tactile clip as $\mathbf X\in\mathbb R^{T\times N\times3}$, where $T$ is the number of frames and $N$ is the total number of valid taxels across fingers. Each taxel is associated with a three-dimensional position and a finger identity. Sensor layouts may differ in taxel count and arrangement, while single-channel observations are zero-padded to the common three-channel interface. We seek a foundational tactile encoder that maps these observations into a shared representation space, $\mathbf Z=E_\theta^{(g)}(\mathbf X;\mathbf A)$, where $g$ selects the sensing-modality branch and $\mathbf A$ denotes the canonical five-dimensional token addresses.

\begin{figure*}[!t]
    \centering
    \includegraphics[width=1.0\linewidth]{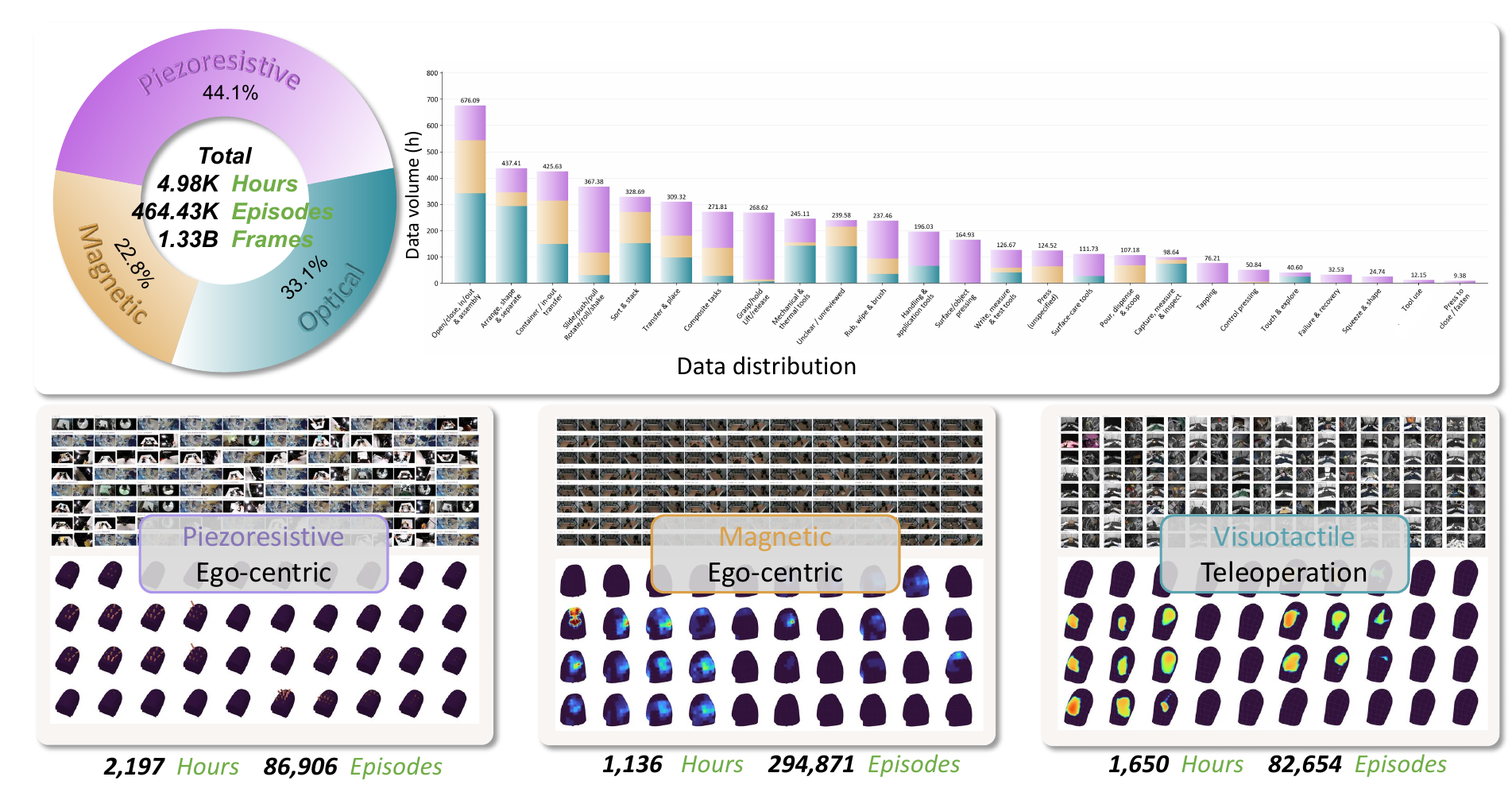}
    \caption{\textbf{Composition and diversity of the LingBot-Touch pretraining dataset.} Our dataset comprises 4{,}983 hours of tactile recordings, spanning 464K episodes and 1.33 billion frames across piezoresistive, magnetic, and visuotactile sensing modalities. To support representation learning for downstream understanding of contact physics and robotic dexterous manipulation, it covers diverse real-world human and robot manipulation tasks, encompassing varied sensing principles, spatial resolutions, sampling frequencies, and physical taxel layouts.}
    \label{pretrain_data}
\end{figure*}
\subsection{Touch at Scale: 5{,}000 Hours from Tactile Gloves and Robot Hands}
\label{sec:method:corpus}
As summarized in \cref{pretrain_data}, our multi-source tactile dataset comprises 4{,}983 hours of recordings, 464{,}431 interaction episodes, and approximately 1.33 billion frames from piezoresistive, magnetic, and visuotactile tactile sensors. Reaching this scale is hard because every hour requires a sensorized hand physically interacting with the world. We do so by combining two ways of producing such data. Human operators wearing tactile gloves generate the bulk of the corpus. A glove needs no robot, so one operator can record hours of everyday manipulation per day in ordinary environments, and the resulting egocentric recordings cover a breadth of objects and behaviors that would be slow to script for a robot. We use piezoresistive and magnetic gloves for these recordings. Teleoperation contributes the remainder: a dual-arm system with Sharpa Wave dexterous hands, whose fingertips carry visuotactile sensors, records the same families of tasks from the hand we later deploy on. Gloves account for two thirds of the corpus and teleoperation for the remaining third.

The two sources are matched on behavior, not on hardware. Every episode is assigned to one of 24 task categories, from opening, closing, and assembly to wiping, pouring, and tool use (\cref{pretrain_data}), and most categories are covered by more than one source. The sensors, by contrast, differ in transduction principle, taxels per finger, sampling rate, and fingertip geometry (\cref{tab:corpus}). Comparable interactions observed through incomparable sensors are what allow a shared encoder to learn regularities of contact that belong to no single sensor, which~\cref{sec:Multi-Source} tests directly.

Annotations stop at the episode-level category. No clip carries labels of force, contact location, or contact timing. During pretraining, we sample clips from the piezoresistive, magnetic, and visuotactile sources with equal sampling probabilities, and all supervision comes from the latent prediction objective in~\cref{sec:method:tjepa}.

\begin{figure*}[t]
    \centering
    \includegraphics[width=1.0\linewidth]{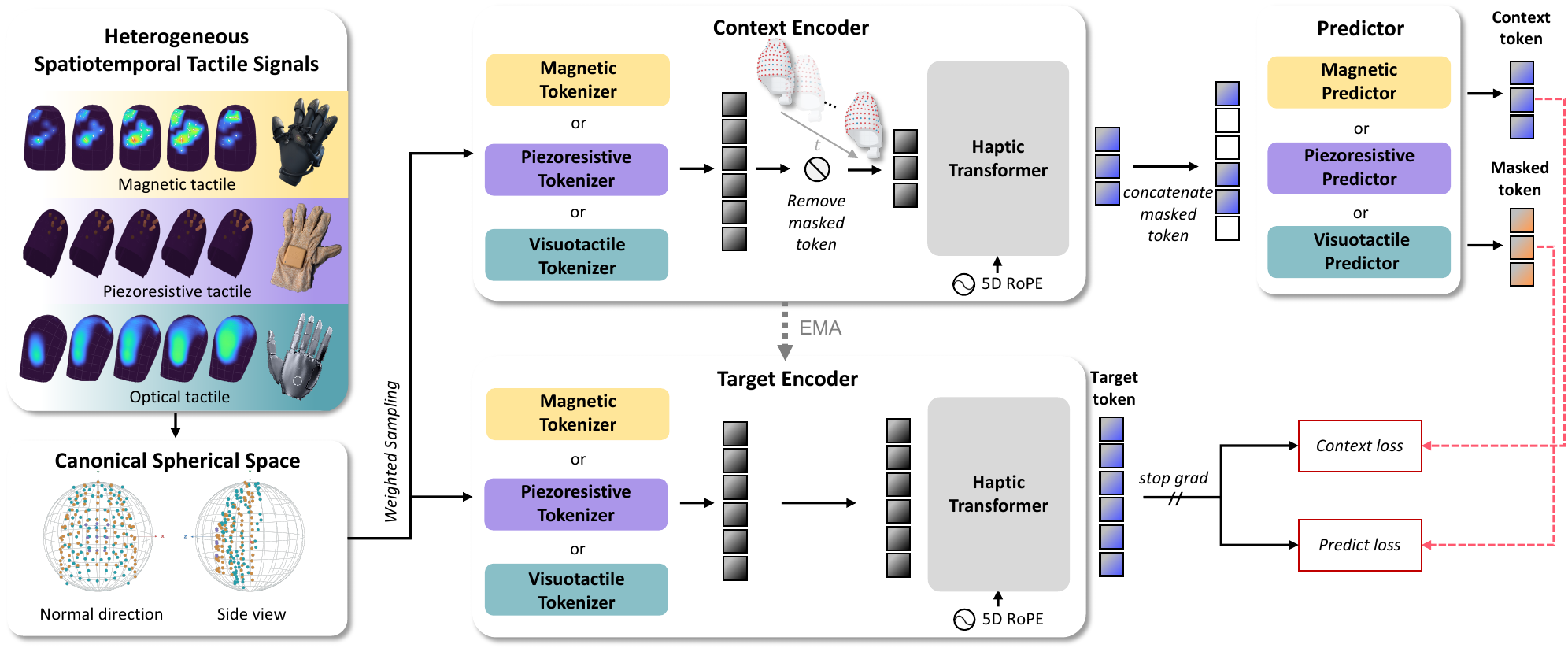}
    \caption{\textbf{Overview of LingBot-Touch.} \method learns unified tactile representations through self-supervised prediction in latent space. Heterogeneous spatiotemporal signals from magnetic, piezoresistive, and visuotactile sensors are mapped to a canonical spherical space and incorporated into pretraining through weighted sampling. Modality-specific tokenizers convert these signals into tokens for a shared Transformer backbone equipped with 5D rotary positional embeddings (RoPE). The context encoder processes only visible tokens, and its outputs are concatenated with mask tokens before being passed to the corresponding modality-specific predictor. The target encoder processes the unmasked input, providing latent targets for the context loss on visible positions and the prediction loss on masked positions. Its parameters are updated through an Exponential Moving Average (EMA) of the context encoder, with gradients stopped through the target branch.}
    \label{method}
\end{figure*}

\subsection{Tactile-JEPA: Predicting Contact in Latent Space}
\label{sec:method:tjepa}
\obj trains the encoder $E_\theta^{(g)}$ by predicting, in latent space, the representation of a complete tactile clip from a partially visible one (\cref{method}). Given a clip $\mathbf X$, we partition its valid positions into a masked set $\mathcal M$ and a visible set $\mathcal C$, and present only $\mathbf X_{\mathcal C}$ to a context encoder. A target encoder processes the complete clip, and a predictor must map the context encoder's output to the target encoder's output at every position, masked or visible. Neither branch reconstructs the signal itself, so the objective is not dominated by the near-zero background that fills most of a tactile clip;~\cref{sec:exp:vae} compares this choice against signal-space reconstruction. Making the scheme work across sensors requires a token address that is meaningful for any sensor, a backbone shared across sensors except at its boundaries, and masking and loss choices suited to sparse signals. We describe each in turn.

\paragraph{Canonical fingertip space and token addresses.}
Taxel indices describe sensor-local ordering but do not identify corresponding
physical locations across sensors. We therefore describe where a measurement
was taken by a continuous 3D coordinate and a finger identity, and place the
coordinates of every sensor in one canonical fingertip space. Starting from
the geometric assets of each sensor, we align the fingertip with a canonical
orientation, then scale and translate it into a shared finger-local coordinate
frame with a unit-radius reference geometry (\cref{method}, bottom left):
\begin{equation}
\mathbf p_i=\alpha\,\tilde{\mathbf p}_i+\mathbf b,
\label{eq:canonical_fingertip}
\end{equation}
where $\tilde{\mathbf p}_i$ is the orientation-aligned position of the taxel
associated with token $i$, and $\alpha>0$ and $\mathbf b\in\mathbb{R}^3$ are
fixed offline for each sensor--finger template. For PaXini and XHand1, we set
$\alpha=1/r$ and $\mathbf b=-\mathbf o/r$, where $\mathbf c$ and $r$ are the
center and radius of the minimum enclosing sphere of the taxel positions
and reference fingertip geometry. The coordinate origin is
$\mathbf o=(u_x,c_y,u_z)$, where $\mathbf u$ is the designated contact-center
anchor and $y$ denotes the finger direction. Sharpa Wave follows the same
height-aligned normalization using its dense surface points, with an
additional depth translation matching the PaXini reference sphere.
Juqiao and Wujie use fixed PaXini reference radii and contact-region anchors.
Tachin uses the maximum isotropic fit within the reference sphere subject
to pulp-depth alignment, followed by a scale factor of $0.999$.
These parameters are reused across temporal windows.

Each valid token then receives a five-dimensional address:
\begin{equation}
\mathbf a_i=(t_i,f_i,\mathbf p_i)\in\mathbb{R}^5,
\qquad
\mathbf A=(\mathbf a_i)_{i=1}^{L},
\label{eq:tactile_address}
\end{equation}
where $t_i$ indexes the token's temporal window, $f_i$ identifies the finger,
and $L$ is the number of valid tokens; a taxel keeps the same $\mathbf p_i$
across windows. The address enters the model only through rotary position
embeddings (RoPE) applied to attention queries and keys, with each of the
five axes acting on its own slice of the head dimension. This positional
encoding does not require a fixed token count or a sensor-specific
positional table, allowing a shared backbone to process different sensor
layouts and supporting transfer to unseen sensors.

Finally, signal amplitudes are normalized by fixed channel response bounds
associated with each sensor's measurement range: nonnegative normal responses
are mapped to $[0,1]$ and signed tangential responses to $[-1,1]$.
Together with the canonical geometry, these dimensionless amplitudes form
the common input interface for multi-source learning.

\paragraph{Modality-specific tokenizers and a shared Haptic Transformer.}
Sensor-specific parameters are confined to shallow projections at the boundaries of the network. We use three tokenizers, one each for magnetic, piezoresistive, and visuotactile sensing. Each is a two-layer MLP that embeds two consecutive frames of a single valid taxel, six numbers in all, into one token, so tokenization preserves the spatial granularity of the sensor and a clip of $T$ frames yields $L=NT/2$ tokens. The tokens then enter the Haptic Transformer (HaT), a Transformer encoder shared across all sources that incorporates time, finger identity, and canonical position through the five-axis RoPE above. The tokenizers extract shallow features from heterogeneous measurements, while successive HaT layers learn shared tactile representations at increasing levels of abstraction. For sensing group $g$, we denote the tokenizer together with HaT by $E_\theta^{(g)}$. The predictor $P_\phi^{(g)}$ follows the same pattern: group-specific input and output projections around a shared prediction Transformer, with the same five-dimensional positional encoding. Predictions and targets are
\begin{equation}
\begin{aligned}
\hat{\mathbf z}
&=
P_\phi^{(g)}
\left(
E_\theta^{(g)}(\mathbf X_{\mathcal C};\mathbf A_{\mathcal C}),
\Delta_{\mathcal M};
\mathbf A
\right),\\
\bar{\mathbf z}
&=
\bar E_{\bar\theta}^{(g)}(\mathbf X;\mathbf A),
\end{aligned}
\label{eq:tactile_jepa_forward}
\end{equation}
where arguments after the semicolon specify positional conditioning through RoPE, and $\Delta_{\mathcal M}$ denotes learnable mask tokens placed at the masked addresses. The context encoder sees visible tokens only; the predictor processes visible-token representations and mask tokens jointly and outputs a prediction for both sets. $\bar E_{\bar\theta}^{(g)}$ is the target encoder, with the multi-level readout described below. Predictions and targets are aligned by their original token indices.

\paragraph{Targets and objective.}
Following the multi-level supervision of V-JEPA~2.1~\cite{vjepa21}, the target encoder processes the complete clip, and we take its features from four depths, normalize each individually, and concatenate them to form $\bar{\mathbf z}$. The context encoder concatenates features from the corresponding layers so that the predictor receives multi-level context. We supervise masked and visible positions alike by minimizing
\begin{equation}
\begin{aligned}
\mathcal L
={}&
\frac{1}{|\mathcal M|D}
\sum_{i\in\mathcal M}
\left\|
\hat{\mathbf z}_i-\operatorname{sg}(\bar{\mathbf z}_i)
\right\|_1
+
\frac{\lambda}{|\mathcal C|D}
\sum_{i\in\mathcal C}
\left\|
\hat{\mathbf z}_i-\operatorname{sg}(\bar{\mathbf z}_i)
\right\|_1,
\end{aligned}
\label{eq:tactile_jepa_loss}
\end{equation}
where $D$ is the concatenated feature dimension, $\operatorname{sg}$ denotes stop-gradient, and $\lambda=0.5$. The prediction loss supervises representations at masked positions, encouraging the model to infer missing contact information from the visible context. The context loss aligns predictions at visible positions with targets computed from the complete clip, encouraging the preservation of observed contact information. The loss is averaged over sampled clips. Gradients update the online tokenizers, HaT, and predictor, including the learnable mask tokens parameterized by $\phi$. The target tokenizers and encoder are updated exclusively through an EMA of their online counterparts.
For downstream tasks, we freeze the pretrained EMA target tokenizers and encoder. Each input is routed to the appropriate tokenizer branch according to its sensor type and processed by the shared HaT backbone. Token features from four selected HaT layers are individually layer-normalized and concatenated along the feature dimension, providing representations at multiple depths for task-specific probes and control policies.

\begin{wrapfigure}{r}{0.54\linewidth}
    \centering
    \includegraphics[width=\linewidth]{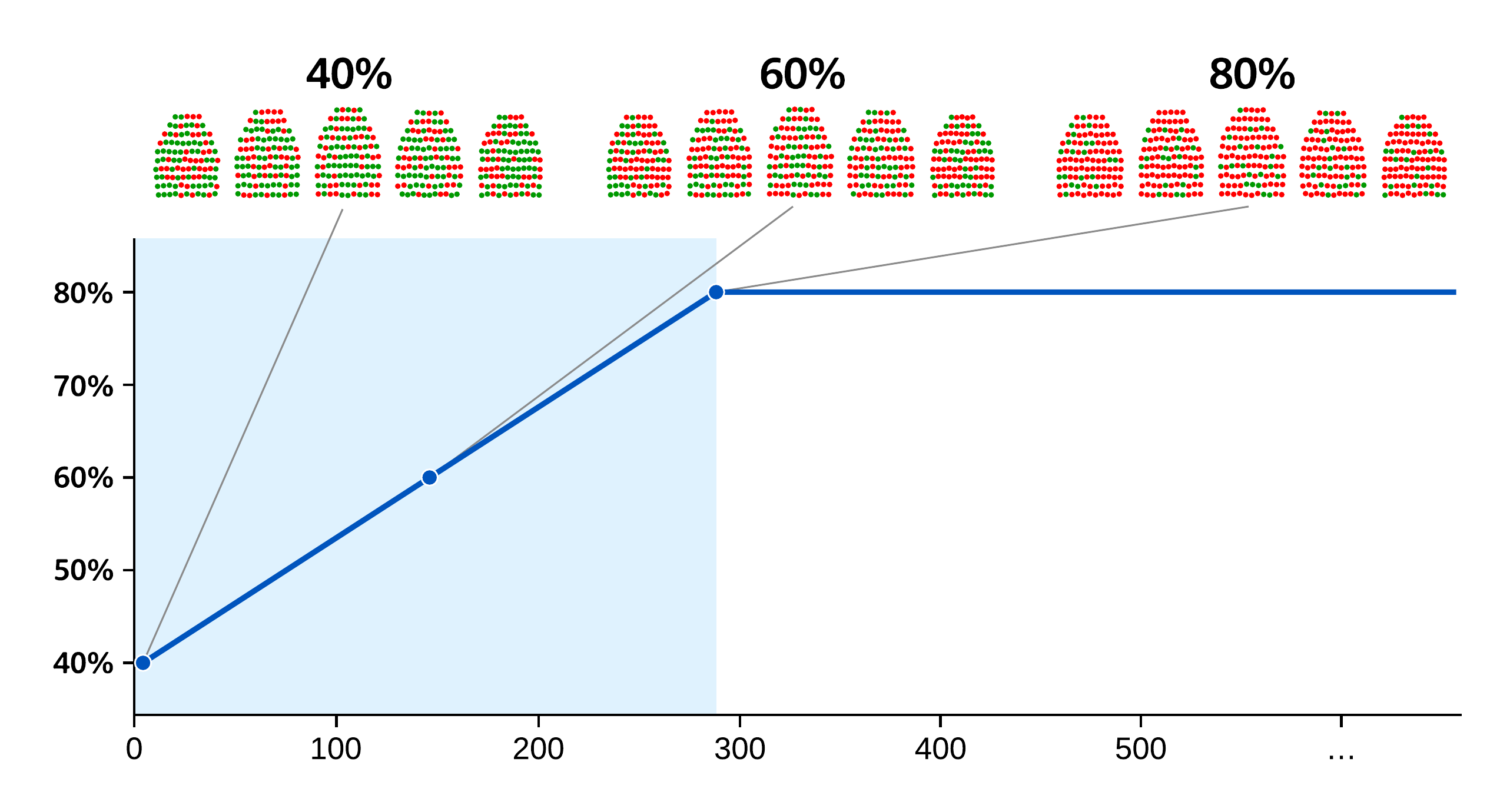}
    \setlength{\abovecaptionskip}{3pt}
    \setlength{\belowcaptionskip}{0pt}
    \caption{\textbf{Spatial mask-ratio warm-up.}
    The masking ratio increases linearly from 40\% to 80\%
    over the first 300 training steps, then remains fixed.
    Red and green denote masked and visible taxels.}
    \label{warmup}
\end{wrapfigure}
\paragraph{Masking and mask warm-up.}
We randomly select valid taxels for masking and extend the same spatial mask throughout the clip, so that $\mathcal M$ contains every token of a masked taxel. The context encoder therefore cannot read a masked taxel's response at another time step and must infer it from the surrounding taxels and their temporal evolution. The final masking ratio is $80\%$ of valid taxels. Tactile signals are sparse, however, and this ratio cannot be applied from the first step. If contact involves $k$ taxels and masking is uniform, all $k$ are hidden with probability about $0.8^k$, so for the few taxels a fingertip typically presses on, a large share of clips present the predictor with a context that contains no contact at all, while its targets, computed from the full clip, do. The loss is then dominated by targets that are unpredictable from the input, and in practice we observe gradient explosions and unstable optimization. We therefore warm up the ratio (\cref{warmup}): it increases linearly from $40\%$ to $80\%$ over the first 300 optimizer steps and stays fixed thereafter. The warm-up spans a negligible fraction of training (we pretrain \method for a total of 200{,}000 optimizer updates), lowers the chance of erasing all contact evidence while the encoder is still untrained, and then raises prediction difficulty to the level at which the rest of pretraining runs.

\subsection{Layout-Agnostic Augmentation}
\label{sec:method:aug}
The canonical fingertip space of~\cref{sec:method:tjepa} gives every sensor's taxels comparable coordinates, but it does not by itself make the encoder indifferent to layout. Pretraining presents exactly three layouts, one per sensor, and RoPE hands the backbone their exact coordinates, so nothing prevents it from memorizing three point sets. A hand it has never seen is a fourth. We therefore treat the layout as a nuisance variable and randomize it during pretraining in two ways: the density at which a fingertip is sampled, and the precise position of each taxel.~\cref{sec:exp:scalingtransfer} tests the result on a sensor absent from pretraining, and~\cref{sec:exp:ablation} ablates the augmentation.

\paragraph{Multiple sampling densities.}
We expose the model to the same interaction at multiple spatial resolutions. For magnetic sensing, we construct views with 24, 32, 48, 64, 76, and 96 taxels per finger from native layouts containing 116 taxels on the thumb and 96 on each remaining finger. The 96-taxel view uses a fixed subset of the thumb taxels and retains all taxels on the remaining fingers; lower-resolution views are obtained by resampling the native measurements. For visuotactile sensing, we aggregate deformation-map responses into 32, 96, or 200 virtual taxels per finger. Resampling uses local Gaussian weights based on the 3D distances between source and target positions and operates on the observed response channels. For each training clip, we uniformly sample one view from the resolutions available for its source.

We retain the native layouts of the piezoresistive sources. In particular, the Juqiao and Xspark fabric gloves provide coarse fingertip arrays of $3\times3$ and $4\times4$ taxels, respectively, making further downsampling undesirable because it would discard already limited spatial detail. Across the multi-resolution views, the underlying interaction remains shared while spatial sampling varies, encouraging representations that generalize across resolutions.

\paragraph{Taxel position perturbation.}
To discourage shortcuts based on exact coordinate templates, we perturb canonical taxel positions by offsets bounded by local spacing:
\begin{equation}
\mathbf p'_i=\mathbf p_i+\boldsymbol\delta_i,\qquad \|\boldsymbol\delta_i\|_2\le 0.5\,d_i,
\label{eq:position_jitter}
\end{equation}
where $\mathbf p_i$ is the canonical coordinate of Eq.~\eqref{eq:canonical_fingertip} and $d_i$ is the distance from taxel $i$ to its nearest other valid taxel on the same finger. The bound scales with local sampling density and keeps every displacement inside the taxel's own neighborhood. Offsets are resampled for each training clip and shared by a taxel across all temporal windows and across the context and target branches, so both branches see the same geometry and the target stays consistent with the context. The perturbed coordinates replace $\mathbf p_i$ in the address, giving $\mathbf a'_i=(t_i,f_i,\mathbf p'_i)$. The response sequence and its temporal correspondence are untouched; only the positional encoding sees a nearby but different layout. We report the data augmentation ablations in \cref{tab:tactile_jepa_ablations_sharpa}. These results show that data augmentation substantially improves \method's performance on downstream tactile understanding tasks.

\section{Experiments: Probing the Tactile Representation}
\label{sec:exp:probe}

This section asks whether \method learns what it was built to learn, and why. Throughout, the pretrained encoder is frozen and read out by lightweight probes, so every number measures the representation rather than a task-specific model. We ask three questions. Does the representation expose the physics of contact, and does it do so on a sensor the model never saw (\cref{sec:ContactPhysics})? How does it change with more hours, more sensor types, and a larger model (\cref{sec:exp:scalingtransfer})? Which design decisions matter, from the latent-prediction objective to masking and augmentation (\cref{sec:exp:vae,sec:exp:ablation})? We close by looking at the latent space itself (\cref{sec:exp:interp}).~\cref{sec:exp:robot} then asks whether the same frozen representation helps robots act.

\subsection{Setup}
\label{sec:exp:setup}

\paragraph{Platforms.}
All probing experiments are conducted on two dexterous hands. Sharpa Wave uses the visuotactile fingertip sensors that provided the robotic portion of the pretraining dataset, so its sensor is seen during pretraining. Each fingertip produces a deformation map at a native resolution of $240\times240$ and a sampling rate of 30 Hz, which we spatially downsample to 96 taxels for model input. XHand1 uses magnetic fingertip sensors of a model absent from the pretraining dataset, differing from the pretraining gloves in spatial resolution, sampling rate, and taxel placement. Each fingertip has 120 native taxels sampled at 30 Hz. In our experiments, we spatially downsample these observations to 96 taxels per finger, although the model can also directly process all 120 native taxels. XHand1 inputs follow the same processing pipeline as other magnetic sources: taxel positions are mapped into the canonical fingertip space using the sensor geometry (\cref{sec:method:tjepa}), and the readings are processed by the magnetic tokenizer. No tokenizer or encoder parameters are updated to accommodate the new sensor.

\paragraph{Models.}
We pretrain two sizes of the Haptic Transformer: \method-Tiny and \method-Small, with 5.46M and 21.76M parameters, respectively. Both use 12 Transformer layers, with embedding dimensions of 192 and 384, respectively. Parameter counts include the tokenizers and shared HaT backbone, excluding the pretraining predictor and downstream task heads. Unless stated otherwise, results in this section use \method-Small pretrained on the full dataset.

\paragraph{Protocol.}
Following the V-JEPA evaluation protocol~\cite{vjepa}, we freeze the EMA target tokenizer and target encoder, read out the encoder's concatenated multi-level features, and train one attentive probe per task: a cross-attention pooling layer followed by a linear head. The tokenizer and encoder receive no task supervision.~\cref{tab:probe_tasks} in \supp lists dataset sizes, inputs, outputs, and metrics for every task, while~\cref{tab:probe_hyperparameters} in \supp provides the detailed probe training budgets and hyperparameters.

\begin{figure*}[t]
    \centering
    \includegraphics[width=1.0\linewidth]{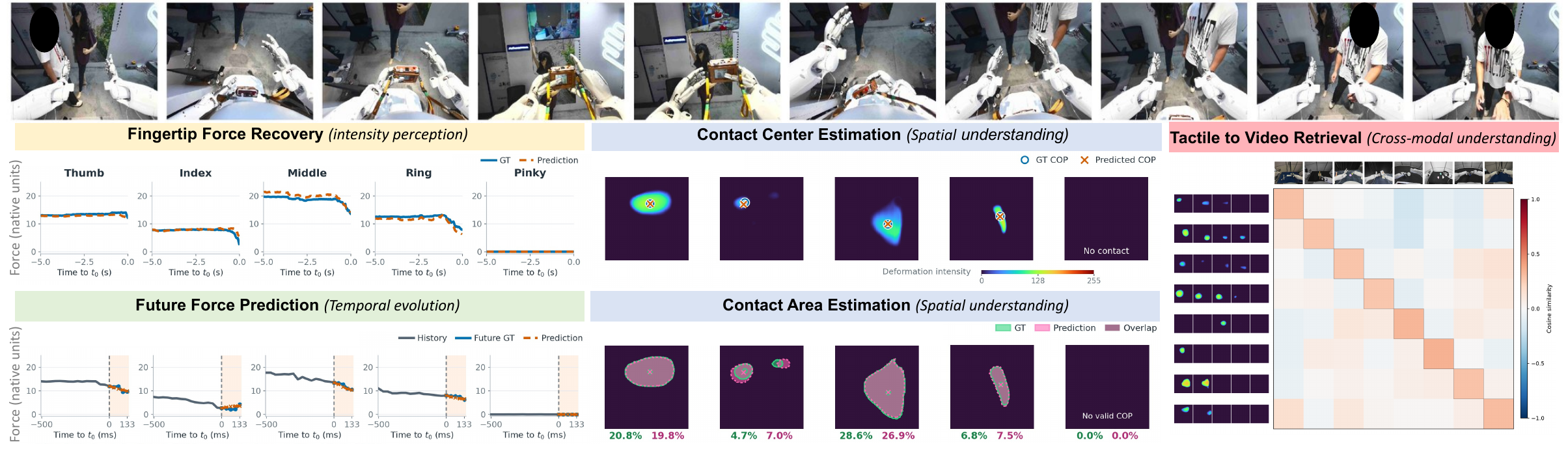}
    \caption{\textbf{Contact-physics probing tasks.} Five frozen-encoder probing tasks on two dexterous hands, Sharpa Wave and XHand1: fingertip force recovery, contact center and coverage estimation, future force prediction, and tactile-to-video retrieval. They test contact intensity, spatial structure, short-term dynamics, and cross-modal correspondence. Each task trains its own attentive probe on the multi-level features of the frozen EMA target encoder. For fingertip force recovery, labels are obtained by averaging the per-finger resultant force vectors over each pair of consecutive frames and then computing their magnitudes. Contact centers are computed as contact-strength-weighted centroids of valid taxels above the activation threshold, while contact coverage is the fraction of valid taxels that are active. Future force labels represent per-finger changes in force magnitude at each of the next four frames relative to the final history frame. Contact centers are supervised only when contact is present.}
    \label{probe_setting}
\end{figure*}

\paragraph{Tasks.}
\textbf{Five tasks} test four aspects of contact (\cref{probe_setting}). For contact intensity, \emph{fingertip force recovery} regresses the resultant contact force in newtons from fingertip signals. For spatial structure, \emph{contact center estimation} localizes the center of pressure, scored by distance error in millimeters, and \emph{contact area estimation} predicts the fraction of the fingertip in contact, scored by coverage error in percentage points. For temporal dynamics, \emph{future force prediction} forecasts the resultant force over the next four frames from the tactile history and the future finger joint states. For cross-modal correspondence, \emph{tactile-to-video retrieval} finds the video frame recorded at the same instant as a tactile clip, scored by top-5 recall. This retrieval task tests whether touch alone identifies the visual context of an interaction.

\subsection{Contact Physics from a Frozen Encoder}
\label{sec:ContactPhysics}
\label{sec:exp:compare}

\begin{figure*}[t]
    \centering
    \includegraphics[width=1.0\linewidth]{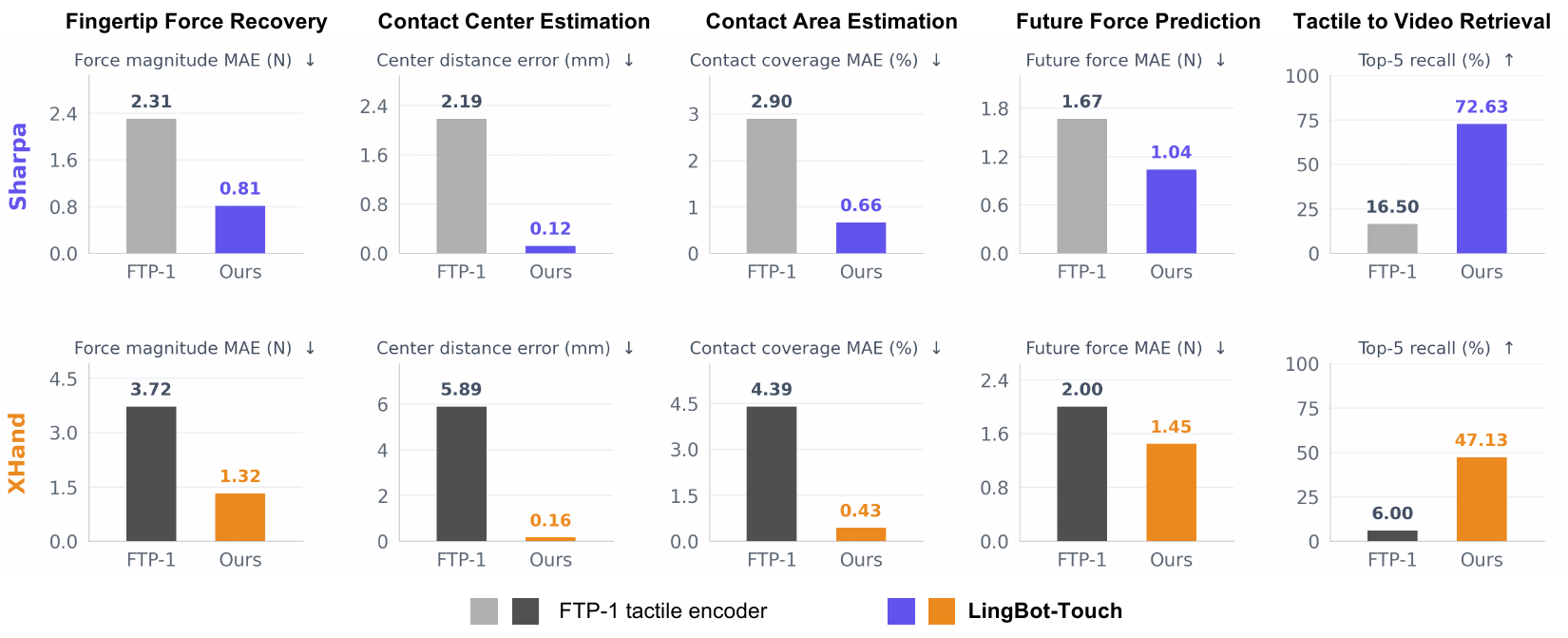}
    \caption{\textbf{Comparison with FTP-1 using frozen pretrained backbones.} Results on Sharpa Wave (top) and XHand1 (bottom). Gray bars show FTP-1 with sensor-specific adapters trained for each platform. Colored bars show \method with the HaT-Small backbone, pretrained on 5K hours of tactile data and evaluated without sensor-specific encoder adaptation. Arrows indicate whether lower or higher values are better.}
    \label{ftp1}
\end{figure*}

\cref{ftp1} reports all five tasks on both hands and compares against FTP-1~\cite{ftp1}, a foundation tactile policy that maps heterogeneous sensors into a morphology-aware token space through sensor-specific encoders. FTP-1 is the closest prior encoder in scope, since it is built for heterogeneous sensors and includes a magnetic branch. For FTP-1, we freeze the released shared tactile backbone and train a separate input adapter for each sensor and probe. The Sharpa Wave
adapter encodes 96 deformation values per finger, replacing the
released image tokenizer, while the XHand1 adapter encodes 120
three-channel taxel measurements per finger. Each adapter is jointly
trained with its probe head for 100 epochs on the corresponding
downstream training split. For \method, we freeze the target
encoder and train only the probe head, without an additional input
adapter.

\begin{figure*}[t]
    \centering
    \includegraphics[width=1.0\linewidth]{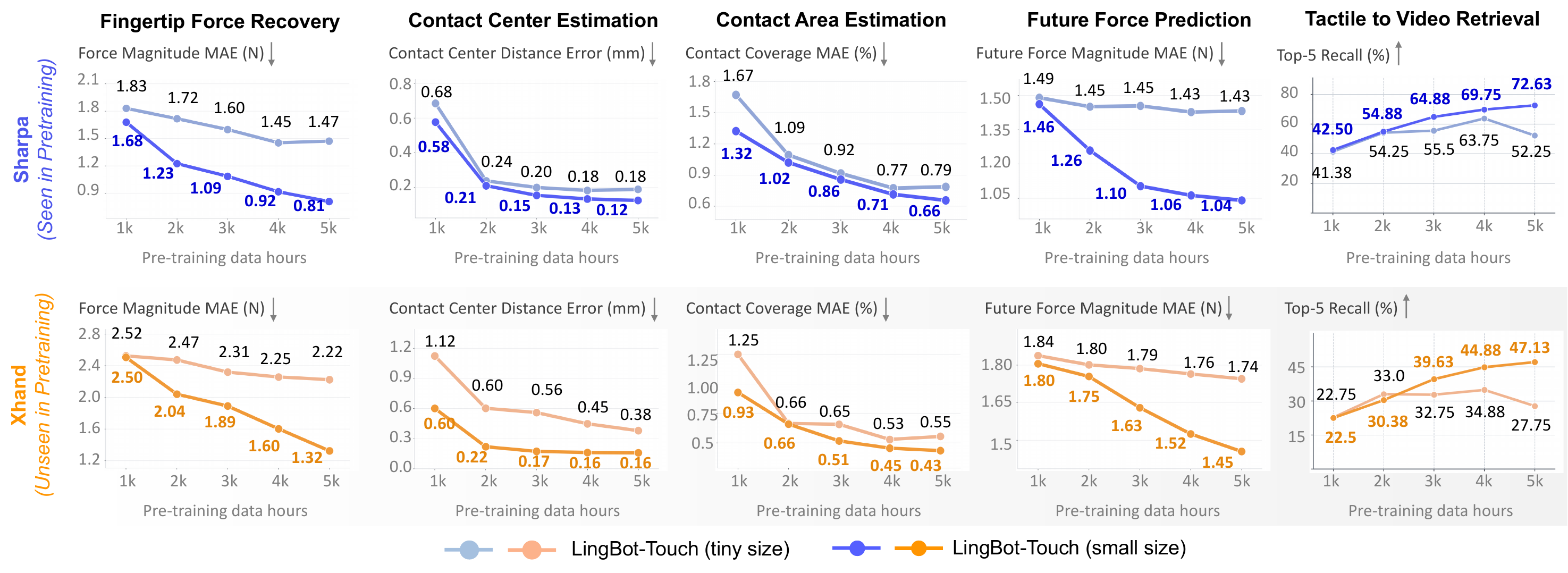}
    \caption{\textbf{Scaling with pretraining data and model size.} Frozen-encoder probing on Sharpa Wave (top) and XHand1 (bottom) for Tiny and Small encoders pretrained on 1K to 5K hours with fixed source proportions. Small improves or holds on every task with every increment of data on both hands; Tiny saturates at about 4K hours and loses retrieval recall at 5K.}
    \label{scaling}
\end{figure*}

\paragraph{On the sensor seen in pretraining.}
On Sharpa Wave, \method recovers fingertip force to within 0.81 N on sensors with a 0 to 30 N range, locates the contact center to within 0.12 mm, and estimates coverage to within 0.66 percentage points, against 2.31 N, 2.19 mm, and 2.90 points for FTP-1. Future force error falls from 1.67 to 1.04 N. The widest gap is in retrieval, where top-5 recall rises from 16.5\% to 72.6\%: the representation carries information about the interaction as a whole, not only about the local contact patch. These margins do not depend on the full corpus. With 1{,}000 hours of pretraining, force and future-force errors are already below 2 N, contact-center error is below 1 mm, and coverage error is below 2 points (\cref{scaling}).

\paragraph{On a sensor never seen.}
On XHand1, with no parameter updated for the sensor, the same encoder reaches 1.32 N, 0.16 mm, 0.43 points, and 1.45 N on the four physical tasks and 47.1\% retrieval recall. FTP-1 fitted with an XHand1-specific adapter reaches 3.72 N, 5.89 mm, 4.39 points, 2.00 N, and 6.0\%. The spatial tasks show the largest margins: contact-center error on the unseen sensor is more than an order of magnitude below FTP-1's error on the seen one. Errors on XHand1 are higher than on Sharpa Wave on four of the five tasks, as expected for a sensor with no pretraining exposure, but the ranking of tasks and the size of the gap to FTP-1 are the same on both hands. This is the behavior the canonical fingertip space and the layout augmentation of~\cref{sec:method:aug} were designed to produce, and it separates \method from encoders that need a dedicated adapter or tokenizer for each sensor model.

\begin{figure*}[!t]
    \centering
    \includegraphics[width=1.0\linewidth]{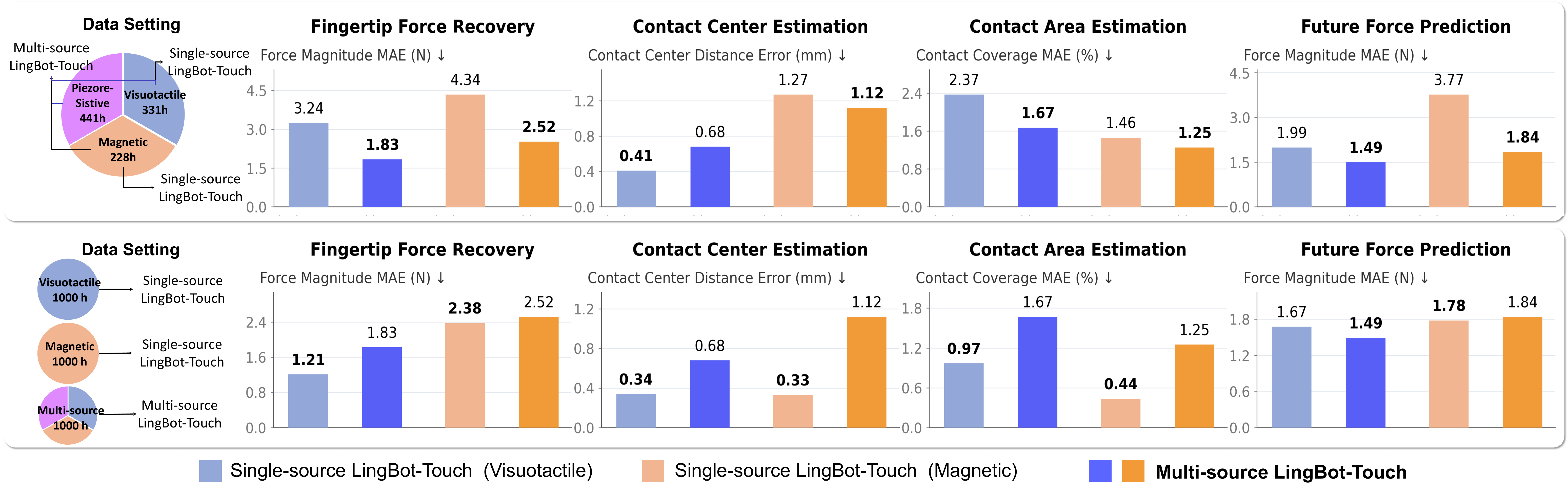}
    \caption{\textbf{Single-source versus multi-source pretraining.} Blue:  probed on Sharpa Wave. Orange: probed on XHand1. Top: adding 441 hours of piezoresistive data and the other sensor's data to a fixed 331 or 228 hours of a sensor's own data improves seven of eight sensor-task pairs. Bottom: at a fixed 1{,}000-hour budget, a single-source model beats the three-source mixture on its own sensor on seven of eight pairs, by the widest margins on the spatial tasks. Bold marks the better of each pair. \method uses the HaT-Tiny backbone.}
    \label{singlevsmulti}
\end{figure*}

\subsection{Scaling the Corpus: Hours, Sensors, and Model Size}
\label{sec:exp:scalingtransfer}
\label{sec:Multi-Source}

The corpus is the lever behind every result in this paper, and it can be turned in three ways: how many hours are used, which sensors they come from, and how large a model absorbs them. We vary each in turn and probe the frozen encoder on both hands.

\paragraph{More hours.}
We pretrain Tiny and Small on five subsets of the corpus, from 1K to 5K hours with the source proportions fixed (\cref{scaling}). Small improves or holds on every task with every additional thousand hours, on both hands; on Sharpa Wave its force error falls from 1.68 to 0.81 N and its retrieval recall rises from 42.5\% to 72.6\%. Tiny tracks the same curve to about 4K hours and then flattens, and at 5K hours its retrieval recall drops on both hands. At that point the data are not the limit; the model is, so capacity has to grow with the corpus. Real-world manipulation improves with pretraining hours in the same way, as~\cref{sec:exp:robot} reports.

\paragraph{Which hours, at fixed per-sensor data.}
The corpus mixes three sensing principles, and we ask whether they help one another (\cref{singlevsmulti}). Probing uses the four physical tasks; single-source visuotactile models are probed on Sharpa Wave and single-source magnetic models on XHand1. We first hold a sensor's own data fixed and add the others: a single-source visuotactile model sees 331 hours and a single-source magnetic model 228 hours, while the multi-source model sees those same hours plus 441 hours of piezoresistive data, 1{,}000 hours in all. The multi-source model improves on seven of the eight sensor-task pairs (\cref{singlevsmulti}, top). On the magnetic side every task improves, force error from 4.34 to 2.52 N and future-force error from 3.77 to 1.84 N; on the visuotactile side force, coverage, and future force improve, while contact-center error rises from 0.41 to 0.68 mm.

\paragraph{Which hours, at fixed total budget.}
We then give each single-source model 1{,}000 hours of its own sensor and compare against the same 1{,}000-hour mixture, in which visuotactile and magnetic data make up 33\% and 23\% (\cref{singlevsmulti}, bottom). Now the single-source model wins on its own sensor on seven of eight pairs: by a wide margin on the spatial tasks (contact center 0.34 versus 0.68 mm for visuotactile, 0.33 versus 1.12 mm for magnetic) and by little on temporal prediction, where the multi-source model is in fact better for visuotactile (1.49 versus 1.67 N). The two settings say the same thing from opposite sides. Force and dynamics transfer across transducers, so hours from another sensor substitute for a sensor's own when those are scarce. Spatial precision depends on seeing a sensor's own layout at its own resolution, so at a fixed budget pooling costs some of it, in exchange for one encoder that serves three sensor families.

\paragraph{The sensor that contributed nothing.}
XHand1 appears in none of these corpora, yet it benefits from every knob. Its probing curves follow the Sharpa Wave curves in shape as hours increase, for both model sizes (\cref{scaling}, bottom), and the magnetic bars of~\cref{singlevsmulti} are XHand1 results: a magnetic model given other sensors' hours gets better on a magnetic sensor it never saw. What improves with more and more varied data is a representation of contact, not of any particular transducer.

\paragraph{Takeaway.}
Pool every sensor, and scale the model with the data. No tactile sensor today has 5{,}000 hours of its own, and these experiments show it does not need them: hours from other transducers carry most of what matters for intensity and dynamics, and the remaining gap in spatial precision is the price of a single encoder that transfers to hands outside the corpus. Further, this reframes what matters when scaling tactile data. Because hours transfer across transducers, the sensor to scale with is not the most capable one but the one at the sweet spot between capability and ease of capture, which is why two thirds of our corpus come from gloves a person can wear all day rather than from the fingertips we later deploy on.

\begin{figure*}[t]
    \centering
    \includegraphics[width=1.0\linewidth]{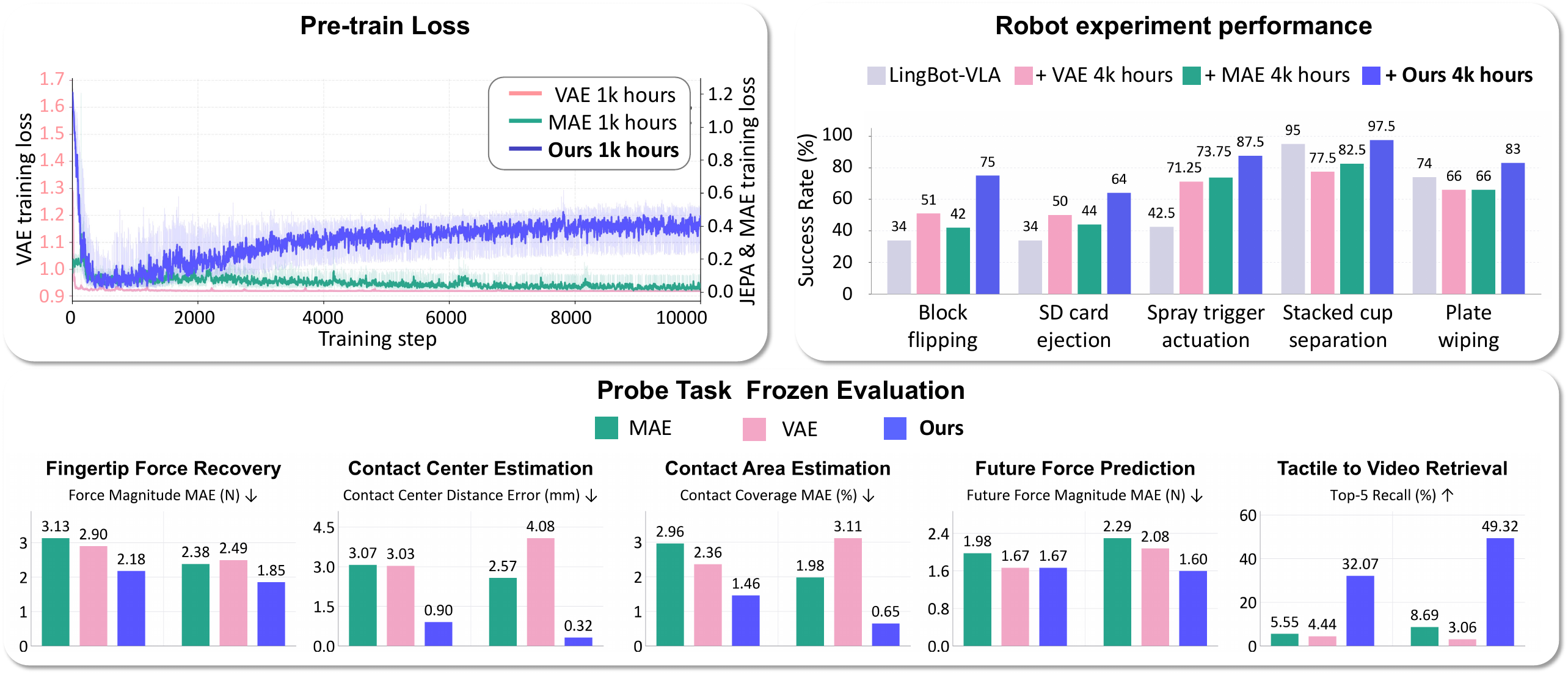}
    \caption{\textbf{Latent prediction versus signal reconstruction.} VAE, MAE, and \method are pretrained on identical data with matched hyperparameters. Top left: pretraining losses. Bottom: frozen-encoder probing with 1K and 4K hours of pretraining data. Top right: real-world manipulation success rates with the pretrained tactile encoders integrated into \lingbotvla.}
    \label{vae}
\end{figure*}

\subsection{Latent Prediction versus Signal Reconstruction}
\label{sec:exp:vae}
\cref{sec:method:tjepa} motivates latent-space prediction for sparse tactile signals. We evaluate this choice by pretraining three models on identical multi-source tactile data with matched hyperparameters: a variational autoencoder (VAE) and a masked autoencoder (MAE), both of which reconstruct signals, and \method. Each model is pretrained at two data scales, 1K and 4K hours, and evaluated on the five probing tasks (\cref{vae}). To ensure comparable model capacity, VAE and MAE each use 5.46M parameters, matching \method with the HaT-Tiny backbone. MAE and \method also use the same masking ratio of 80\% in this comparison.

\paragraph{Training dynamics.}
The VAE loss drops to a low plateau within a few hundred steps and stays there; the MAE loss converges to a small reconstruction error. Neither optimization signal corresponds to a good representation. \method shows a different trajectory: the loss falls, then rises toward a nonzero plateau as the EMA targets evolve, the expected signature of a target that keeps moving. Because the three losses live in different spaces, we compare the representations only through probing.

\paragraph{Probing.}
\method is best or tied-best on every task at both scales. At 1K hours it cuts the contact-center error of either reconstruction model to less than a third (0.90 mm against 3.03 for the VAE and 3.07 for the MAE) and reaches 32.1\% retrieval recall against 4.4\% and 5.6\%. At 4K hours it improves on all five tasks, to 0.32 mm and 49.3\% recall. The reconstruction objectives do not scale the same way. The VAE gets worse with more data on four of five tasks, its contact-center error rising from 3.03 to 4.08 mm and its recall falling from 4.4\% to 3.1\%; the MAE improves on four tasks but its recall stays below 9\%. Reconstructing a signal that is mostly zeros rewards modeling the background, and more background does not help. Predicting in latent space never sees the background as a target.

\begin{table*}[!t]
  \centering
  \caption{\textbf{LingBot-Touch ablations on Sharpa Wave.} We evaluate frozen Tiny encoders pretrained on a 1K-hour data subset for 100 epochs. Unless varied within a block, the default configuration uses 90\% spatial masking, a temporal masking range of $[75,75]\%$, value augmentation only, 4-frame clips, 1-frame tubelets, constant EMA momentum of 0.99925, shared tokenizer and predictor interfaces, and joint target and context losses. Bold numerical entries indicate the best result within each block. The default configuration serves as a common baseline for fair comparison and does not represent our final, best-performing configuration.}
  \label{tab:tactile_jepa_ablations_sharpa}
  \begingroup
  \footnotesize
  \setlength{\tabcolsep}{4pt}
  \renewcommand{\arraystretch}{1.13}
  \begin{tabularx}{\textwidth}{@{}>{\raggedright\arraybackslash}p{2.8cm} >{\raggedright\arraybackslash}p{3.8cm} *{5}{>{\centering\arraybackslash}X}}
    \toprule
    & & \multicolumn{5}{c}{\textit{Sharpa Wave: frozen-encoder evaluation}} \\
    \cmidrule(lr){3-7}
    Ablation & Setting & \shortstack{Force\\recovery MAE\\(N) $\downarrow$} & \shortstack{Center error\\Mean dist.\\(mm) $\downarrow$} & \shortstack{Contact\\coverage\\MAE (pp) $\downarrow$} & \shortstack{Future force\\MAE\\(N) $\downarrow$} & \shortstack{Retrieval\\Recall@5\\(\%) $\uparrow$} \\
    \midrule
    \multirow{5}{*}{\shortstack[l]{Spatial masking\\(\%)}} & 20 & 2.074 & 1.168 & 2.173 & 1.522 & 23.875 \\
     & 40 & 2.189 & 0.945 & 2.328 & 1.526 & 25.375 \\
     & 60 & 2.574 & 1.762 & 3.671 & 1.567 & 24.500 \\
     & 80 & \textbf{1.688} & \textbf{0.223} & \textbf{0.873} & \textbf{1.452} & \textbf{35.375} \\
     & 90 & 2.284 & 2.324 & 3.429 & 1.624 & 14.375 \\
    \midrule
    \multirow{4}{*}{\shortstack[l]{Temporal masking\\(\% of clip)}} & [50,50] & 2.576 & 0.909 & 2.207 & 1.552 & 20.000 \\
     & [75,75] & 2.284 & 2.324 & 3.429 & 1.624 & 14.375 \\
     & [75,100] & 2.155 & \textbf{0.728} & 2.017 & 1.556 & 22.750 \\
     & [100,100] & \textbf{1.886} & 0.904 & \textbf{1.305} & \textbf{1.551} & \textbf{26.500} \\
    \midrule
    \multirow{4}{*}{Data augmentation} & Value + position & \textbf{1.590} & \textbf{0.239} & \textbf{0.931} & 1.478 & \textbf{37.125} \\
     & Value only & 2.284 & 2.324 & 3.429 & 1.624 & 14.375 \\
     & Position only & 1.702 & 0.305 & 0.988 & \textbf{1.453} & 30.625 \\
     & None & 1.808 & 0.639 & 1.428 & 1.508 & 25.625 \\
    \midrule
    \multirow{4}{*}{\shortstack[l]{Clip length\\(frames)}} & 4 & 2.284 & 2.324 & 3.429 & 1.624 & 14.375 \\
     & 8 & 2.383 & 2.407 & 4.519 & 1.655 & 9.500 \\
     & 16 & 1.806 & 0.689 & 1.332 & 1.529 & 29.250 \\
     & 32 & \textbf{1.538} & \textbf{0.207} & \textbf{0.765} & \textbf{1.480} & \textbf{33.250} \\
    \midrule
    \multirow{4}{*}{\shortstack[l]{Tubelet size\\(frames/token)}} & 1 & 2.284 & 2.324 & 3.429 & 1.624 & 14.375 \\
     & 2 & 1.653 & 0.212 & \textbf{0.889} & \textbf{1.400} & 33.375 \\
     & 4 & \textbf{1.496} & \textbf{0.182} & 1.027 & 1.446 & \textbf{35.125} \\
     & 8 & 2.931 & 2.193 & 4.218 & 1.648 & 13.375 \\
    \midrule
    \multirow{3}{*}{EMA momentum} & $0.998\rightarrow1.0$ & 7.221 & 3.401 & 8.221 & 1.728 & 1.375 \\
     & $0.999\rightarrow1.0$ & \textbf{1.529} & \textbf{0.271} & \textbf{1.361} & \textbf{1.463} & \textbf{34.625} \\
     & 0.99925 (constant) & 2.284 & 2.324 & 3.429 & 1.624 & 14.375 \\
    \midrule
    \multirow{2}{*}{\shortstack[l]{Tokenizer /\\predictor interfaces}} & Shared / shared & 2.284 & 2.324 & 3.429 & 1.624 & 14.375 \\
     & Grouped / group-routed & \textbf{1.682} & \textbf{0.702} & \textbf{1.147} & \textbf{1.562} & \textbf{21.750} \\
    \midrule
    \multirow{2}{*}{Training objective} & Target + context & \textbf{2.284} & \textbf{2.324} & 3.429 & 1.624 & \textbf{14.375} \\
     & Target only & 2.303 & 2.380 & \textbf{3.137} & \textbf{1.610} & 14.000 \\
    \bottomrule
  \end{tabularx}
  \endgroup
\end{table*}

\subsection{Ablation Study}
\label{sec:exp:ablation}

\cref{tab:tactile_jepa_ablations_sharpa} varies one design choice at a time and probes the frozen encoder on the five Sharpa Wave tasks. All variants use a Tiny encoder pretrained for 100 epochs on the 1K-hour subset, so absolute numbers sit below those of the full models and only comparisons within a block are meaningful. 

\paragraph{Masking.}
An 80\% spatial mask is best on all five metrics; against 90\% it lowers force error from 2.28 to 1.69 N and raises retrieval recall from 14.4\% to 35.4\%. V-JEPA reports that higher ratios yield more abstract features; for sparse tactile signals the ratio must instead be balanced against how much contact evidence survives in the context, the same tension that motivates the mask warm-up of~\cref{sec:method:tjepa}. Extending the mask across the whole clip beats shorter temporal spans on four of five metrics, confirming that the predictor should not be allowed to copy a masked taxel from a neighboring time step.

\paragraph{Clip length and tubelet size.}
For visuotactile signals at 30 Hz, 32-frame clips (about 1.07 s) with two or four frames per token (67 or 133 ms) are best; single-frame tokens and 8-frame tubelets are markedly worse. These values are tied to the sampling rate and do not transfer as frame counts to sensors running at other rates.

\paragraph{Augmentation.}
Value and position augmentation together are best on four of five metrics, raising retrieval recall from 25.6\% without augmentation to 37.1\%. Position augmentation alone is marginally better on future force (1.45 versus 1.48 N).

\paragraph{EMA momentum.}
A schedule from 0.999 to 1.0 beats both a schedule from 0.998 and a constant 0.99925 on all five metrics. The 0.998 schedule collapses, with force error above 7 N and recall near 1\%, which shows how sensitive latent-target training on sparse signals is to how fast the target moves.

\paragraph{Interfaces and objective.}
Grouped tokenizers with group-routed predictor interfaces beat fully shared interfaces on all five metrics, cutting contact-center error from 2.32 to 0.70 mm; the sensor-specific boundary layers of~\cref{sec:method:tjepa} earn their place. Adding the context loss to the target loss is close to neutral: three metrics improve slightly and two degrade slightly.

\begin{figure*}[t]
    \centering
    \includegraphics[width=1.0\linewidth]{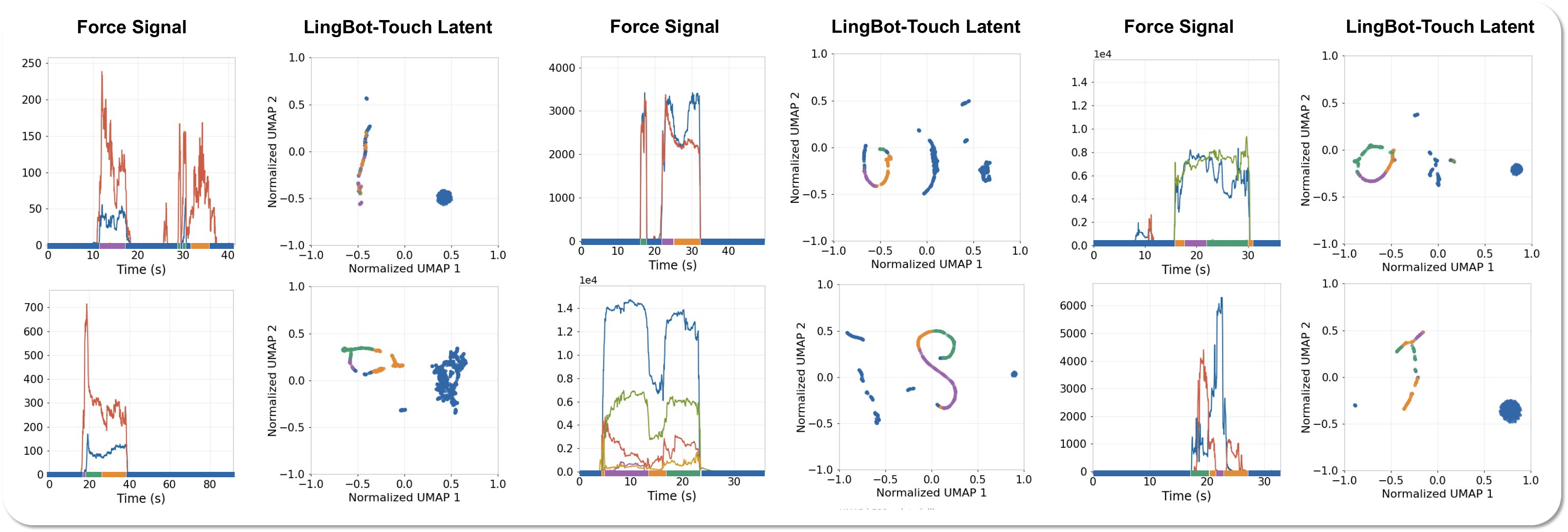}
    \caption{\textbf{Contact separation and continuity in the latent space.} Each example pairs finger-wise tactile responses with phase annotations (left) and a UMAP projection of frozen \method features (right). Each point is a 16-frame window of five-finger observations; point colors follow the phase annotations and curve colors identify fingers. Non-contact windows form one cluster, and contact windows form continuous structures that track the strengthening and weakening of the responses.}
    \label{latent}
\end{figure*}

\subsection{What the Latent Space Encodes}
\label{sec:exp:interp}

We close by looking at how \method organizes short tactile histories, using the phases of robot manipulation as the lens.

\paragraph{Protocol.}
We use the Tiny variant of \method pretrained on the full pretraining dataset. Features are extracted from the frozen target encoder using 16-frame windows with a stride of two frames. We average token features within each finger over time and taxels, then concatenate the five 768-dimensional finger features into a 3{,}840-dimensional representation per window. These representations are projected into two dimensions using UMAP. Each point is colored according to the annotated interaction phase of its corresponding window: non-contact, contact onset, increasing contact force, decreasing contact force, or contact release.

\paragraph{Observations.}
In every example of~\cref{latent}, windows without contact collapse into one tight cluster away from everything else. That part is expected, since their inputs are near zero; what matters is the structure within contact. Contact windows form smooth, extended curves ordered in time: as finger responses strengthen, weaken, and redistribute across fingers, consecutive windows occupy neighboring positions along the curve, and distinct phases of the same interaction occupy distinct segments of it. The representation therefore preserves continuous variation within a contact episode rather than only the contact versus no-contact distinction, which is what a policy needs in order to time its actions.

\section{Experiments: Robot Learning with Frozen Tactile Tokens}
\label{sec:exp:robot}

\cref{sec:exp:probe} showed that the representation exposes contact physics to a probe. This section asks whether a robot can use it. Throughout, \method is a drop-in module: the encoder stays frozen, its tokens are appended to whatever the policy already observes, and nothing else about the policy or its training changes. We ask three questions. Does adding the tokens improve imitation-learning and VLA policies on real hardware, and by how much more than adding the raw tactile signal (\cref{sec:exp:robot:real})? Does it improve reinforcement learning in simulation, in final performance and in the stability of training (\cref{sec:exp:robot:sim})? And does representation quality, as measured by probing, carry over to the policy when we vary the pretraining data or swap in another tactile encoder (\cref{sec:exp:robot:which})?

\subsection{Setup}
\label{sec:exp:robot:setup}

\paragraph{Platforms.}
The same two hands as in~\cref{sec:exp:probe}, each mounted on a dual-arm system: Sharpa Wave, whose visuotactile sensors are seen in pretraining, and XHand1, whose magnetic sensors are not. Both hands are also simulated for the reinforcement-learning experiments, with tactile signals rendered as deformation maps from geometric penetration depth (\cref{app:simulation}).

\paragraph{Integration.}
On the real robots a frozen \method-Tiny encoder, 5.46M parameters, encodes 16-frame tactile windows sampled at 30 Hz for each active hand. Trainable cross-attention readouts aggregate its outputs into 40 tokens per hand, which are linearly projected to each policy's conditioning dimension and enter the policy as an additional conditioning input beside vision and proprioception. The encoder adds 0.16\%, 0.15\%, and 0.13\% to the parameter counts of T-Rex, $\pi_{0.5}$, and \lingbotvla. In simulation, the same frozen encoder processes eight-frame tactile windows, and trainable cross-attention readouts compress its tokens into features concatenated with non-tactile observations as policy inputs (\cref{app:sim_rl_settings}).

\paragraph{Real-world protocol.}
We post-train four policies, Diffusion Policy as an imitation-learning baseline and $\pi_{0.5}$, \lingbotvla, and T-Rex as VLA baselines~\cite{chi2023diffusionpolicy, pmlr-v305-black25a, wu2026foundation, niu2026trextactilereactivedexterousmanipulation}, in three configurations each: the original policy without tactile input, the policy with the raw tactile signal added, and the policy with \method tokens added. All three see the same demonstrations, 200 episodes per task and 500 for tasks that need both hands, and each is evaluated over 20 consecutive trials per task.~\cref{app:real_robot_tasks} describes every task and its success criterion.

\paragraph{Simulation protocol.}
Twelve dexterous tasks, eight on Sharpa Wave and four on XHand1, follow prior tactile reinforcement-learning suites~\cite{miller2025tactilerl,pan2026binarysimtorealdexterousmanipulation}: Baoding rotation, in-hand rotation, nut assembly, plate spinning, pen spinning, blind grasping, grasp stabilization, and ball balancing on Sharpa Wave, and Baoding rotation, ball balancing, blind grasping, and grasp stabilization on XHand1. Policies are trained with PPO over three seeds and compared on success rate and return under five tactile observations: none, a binary contact indicator per taxel, the raw signal, features from a CNN trained with the policy, and frozen \method features.

\subsection{Real-World Manipulation with Imitation Learning and VLA Policies}
\label{sec:exp:robot:real}

\begin{figure*}[!t]
    \centering
    \includegraphics[width=0.7\linewidth]{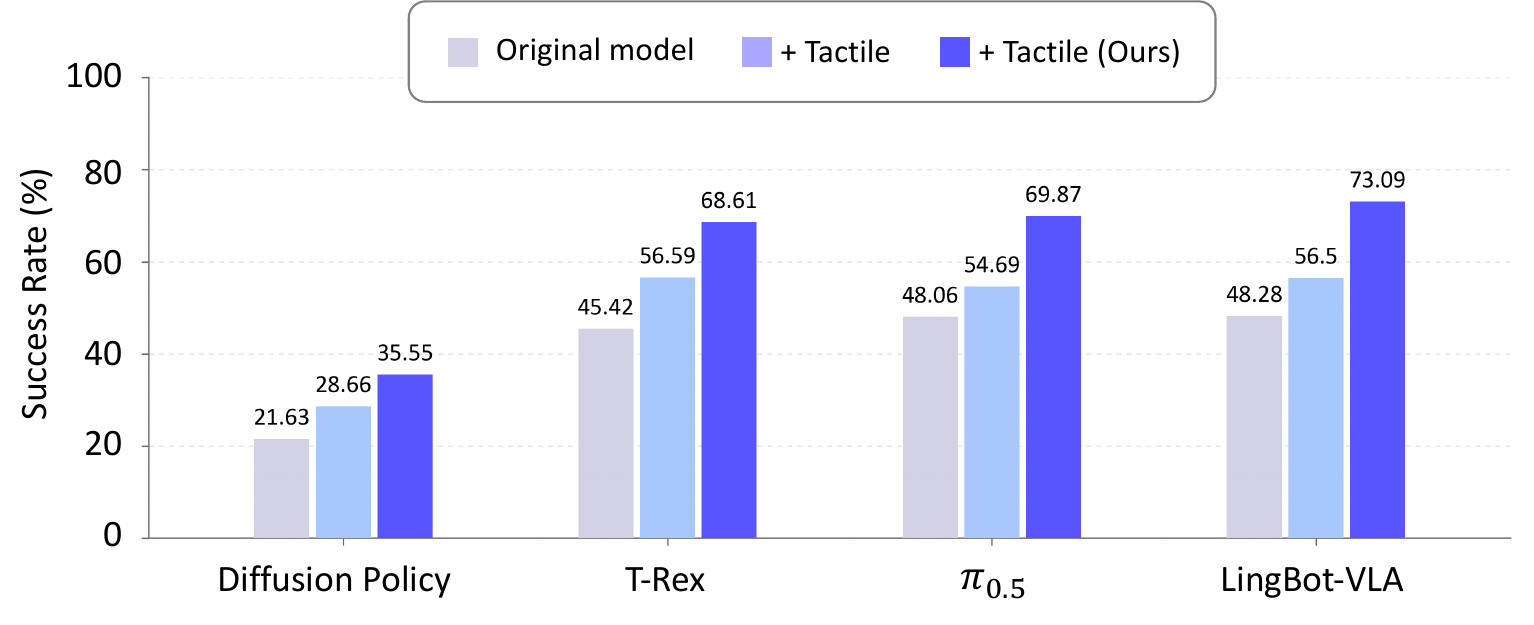}
    \caption{\textbf{Mean success over 26 real-world tasks for four policies.} \method enters only at post-training, as frozen tokens conditioning the policy; it plays no part in any baseline's pretraining. The raw-tactile configuration is the control.}
    \label{realexperiment}
\end{figure*}

\paragraph{Tasks.}
The 26 tasks, 20 on Sharpa Wave and 6 on XHand1, are chosen so that touch matters. They include fine manipulation (SD card ejection, pen uncapping, straw insertion), contact-rich actions (cap twisting, spray trigger actuation, zipper closing), fragile and deformable objects (egg transfer, deformable cup transport, bag extraction), tactile discrimination (stiffness sorting, mahjong sorting), and bimanual coordination (box handover, book extraction).~\cref{tab:real_robot} lists every task and~\cref{realexperiment} the means.

\paragraph{Across policies.}
Averaged over the 26 tasks, \method tokens raise the success rate of Diffusion Policy from 21.6\% to 35.6\%, of $\pi_{0.5}$ from 48.1\% to 69.9\%, of \lingbotvla from 48.3\% to 73.1\%, and of T-Rex from 45.4\% to 68.6\%: gains of 13.9, 21.8, 24.8, and 23.2 percentage points for well under a fifth of a percent in parameters. The gain is not confined to a few tasks. \method is the best of the three configurations in 89 of the 104 task-policy pairs and at or above the original policy in 99 of them.

\paragraph{Representation, not modality.}
The raw-tactile configuration is the control that separates having touch from understanding it. Adding the raw signal helps about half as much at best, 7.0, 6.6, 8.2, and 11.2 points for the four policies, and it hurts in 22 of the 104 pairs, in some cases badly: \lingbotvla falls from 28.8\% to 5.0\% on mahjong sorting and from 65.0\% to 28.8\% on egg transfer, and $\pi_{0.5}$ from 39.0\% to 15.0\% on zipper closing. Given 200 demonstrations, a policy cannot learn to read a sparse, high-dimensional sensor from scratch, and the attempt can distract it from the signals it already reads. The frozen tokens arrive already meaning something.

\paragraph{On the hand the encoder never saw.}
XHand1 makes the point sharply. On its six tasks the raw signal is nearly useless to the policies, changing the mean success of Diffusion Policy, $\pi_{0.5}$, \lingbotvla, and T-Rex by $+4.6$, $-1.7$, $0.0$, and $+2.0$ points, while \method tokens computed from the same signal, with no sensor-specific training, add 16.3, 24.5, 17.8, and 13.8 points. Whatever makes an unfamiliar sensor hard for a policy to read, the canonical representation of~\cref{sec:method:tjepa} removes it.

\paragraph{Where the gains come from.}
The largest per-task gains sit at the step of a task that vision cannot resolve: whether the zipper slider is engaged (\lingbotvla, 11\% to 55\%), whether the trigger has been pulled far enough (\lingbotvla, 42.5\% to 87.5\%), whether the card has released (T-Rex, 32\% to 92\%), whether the cap has broken loose (T-Rex, 56\% to 90\%), and whether the top cup has separated from the stack ($\pi_{0.5}$ on XHand1, 32.5\% to 72.5\%). In the recorded runs the tactile-conditioned policies adjust grip and timing at exactly those moments instead of replaying the demonstrated trajectory. \method is not uniformly better: it falls below the raw-signal configuration in 9 of the 104 pairs, five of them for Diffusion Policy, and below the original policy in 5.~\cref{app:real_robot_tasks} analyzes the behaviors task by task.

\begin{table*}[!t]
    \centering
    \caption{\textbf{Real-robot success rates on Sharpa Wave and XHand1.} Success rates are reported in percent over 20 trials per task. For each policy, we compare the original model, the model augmented with raw tactile signals, and the model augmented with tokens from a frozen \method{} encoder (Ours). We use the HaT-Tiny variant of \method{}, pretrained on 4K hours of data. Means weight all 20 Sharpa Wave tasks or all 6 XHand1 tasks equally; bold indicates the highest mean within each policy for each hand.}
    \label{tab:real_robot}
    \footnotesize
    \setlength{\tabcolsep}{2pt}
    \renewcommand{\arraystretch}{1.0}
\begin{tabular}{@{}cl@{\hspace{2pt}}
                rrr@{\hspace{8pt}}
                rrr@{\hspace{8pt}}
                rrr@{\hspace{8pt}}
                rrr@{}}
                \toprule
                & &
                \multicolumn{3}{c}{Diffusion Policy} &
                \multicolumn{3}{c}{$\pi_{0.5}$} &
                \multicolumn{3}{c}{\lingbotvla} &
                \multicolumn{3}{c}{T-Rex} \\
                \cmidrule(lr){3-5}
                \cmidrule(lr){6-8}
                \cmidrule(lr){9-11}
                \cmidrule(lr){12-14}
                Hand & Task &
                Baseline & + Tactile & + Ours &
                Baseline & + Tactile & + Ours &
                Baseline & + Tactile & + Ours &
                Baseline & + Tactile & + Ours \\
                \midrule

                \multirow{21}{*}{\rotatebox[origin=c]{90}{Sharpa Wave}}
                & Single-page turning
                & 3.75 & 21.25 & 30.00
                & 30.00 & 76.25 & 86.25
                & 23.75 & 60.00 & 67.50
                & 52.50 & 55.00 & 67.50 \\

                & Plate wiping
                & 24.00 & 44.00 & 47.00
                & 68.00 & 74.00 & 76.00
                & 74.00 & 81.00 & 83.00
                & 50.00 & 75.00 & 75.00 \\

                & Stacked cup separation
                & 50.00 & 45.00 & 70.00
                & 61.25 & 67.50 & 72.50
                & 95.00 & 70.00 & 97.50
                & 42.50 & 65.00 & 80.00 \\

                & Mahjong sorting
                & 22.50 & 40.00 & 28.75
                & 35.00 & 40.00 & 70.00
                & 28.75 & 5.00 & 56.25
                & 35.00 & 47.50 & 58.75 \\

                & Pillbox refilling
                & 8.75 & 12.50 & 12.50
                & 22.50 & 22.50 & 32.50
                & 22.50 & 25.00 & 31.25
                & 26.25 & 23.75 & 41.25 \\

                & Business card extraction
                & 20.00 & 32.50 & 56.25
                & 50.00 & 50.00 & 75.00
                & 42.50 & 61.25 & 75.00
                & 65.00 & 57.50 & 75.00 \\

                & Card dealing
                & 46.25 & 60.00 & 75.00
                & 56.25 & 75.00 & 78.75
                & 58.75 & 65.00 & 75.00
                & 73.75 & 78.75 & 88.75 \\

                & Deformable cup transport
                & 46.25 & 42.50 & 63.75
                & 72.50 & 81.25 & 78.75
                & 93.30 & 93.75 & 96.70
                & 95.00 & 97.50 & 100.00 \\

                & SD card ejection
                & 6.00 & 10.00 & 16.00
                & 47.00 & 69.00 & 64.00
                & 34.00 & 72.00 & 64.00
                & 32.00 & 84.00 & 92.00 \\

                & Block flipping
                & 9.00 & 8.00 & 10.00
                & 67.00 & 84.00 & 90.00
                & 34.00 & 48.00 & 75.00
                & 29.00 & 73.00 & 83.00 \\

                & Pen uncapping
                & 0.00 & 0.00 & 0.00
                & 36.00 & 52.00 & 59.00
                & 33.00 & 48.00 & 65.00
                & 22.00 & 33.00 & 61.00 \\

                & Cap twisting
                & 56.25 & 70.00 & 76.25
                & 78.75 & 85.00 & 91.25
                & 75.00 & 75.00 & 87.50
                & 56.25 & 62.50 & 90.00 \\

                & Bag extraction
                & 11.00 & 20.00 & 16.00
                & 47.50 & 40.00 & 53.75
                & 46.25 & 60.50 & 82.50
                & 40.00 & 46.25 & 62.50 \\

                & Straw insertion
                & 14.00 & 18.00 & 21.00
                & 30.00 & 30.00 & 41.00
                & 35.00 & 56.00 & 59.00
                & 29.00 & 44.00 & 44.00 \\

                & Fan opening
                & 17.50 & 27.50 & 41.25
                & 25.00 & 45.00 & 52.25
                & 31.25 & 50.00 & 70.00
                & 58.75 & 51.25 & 66.25 \\

                & Zipper closing
                & 14.00 & 9.00 & 22.00
                & 39.00 & 15.00 & 59.00
                & 11.00 & 16.00 & 55.00
                & 15.00 & 17.00 & 42.00 \\

                & Spray trigger actuation
                & 30.00 & 32.50 & 28.75
                & 51.25 & 60.00 & 71.25
                & 42.50 & 60.00 & 87.50
                & 28.75 & 65.00 & 76.25 \\

                & Stiffness sorting
                & 48.00 & 52.50 & 28.75
                & 41.25 & 57.50 & 81.25
                & 37.50 & 56.25 & 72.50
                & 32.50 & 63.75 & 68.75 \\

                & Book extraction
                & 11.00 & 38.00 & 41.00
                & 36.00 & 62.00 & 76.00
                & 36.00 & 64.00 & 81.00
                & 50.00 & 69.00 & 59.00 \\

                & Box handover
                & 8.00 & 18.00 & 26.00
                & 57.00 & 48.00 & 63.00
                & 58.00 & 59.00 & 69.00
                & 48.00 & 51.00 & 70.00 \\

                \cmidrule(l){2-14}
                & Mean success rate (\%)
                & 22.31 & 30.06 & \textbf{35.51}
                & 47.56 & 56.70 & \textbf{68.58}
                & 45.60 & 56.29 & \textbf{72.51}
                & 44.06 & 57.99 & \textbf{70.05} \\
                \midrule

                \multirow{7}{*}{\rotatebox[origin=c]{90}{XHand1}}
                & Single-page turning
                & 11.25 & 35.00 & 48.75
                & 53.75 & 50.00 & 91.25
                & 65.00 & 67.50 & 68.75
                & 47.50 & 50.00 & 62.50 \\

                & Egg transfer
                & 32.50 & 25.00 & 22.50
                & 67.50 & 65.00 & 83.75
                & 65.00 & 28.75 & 66.25
                & 55.00 & 33.75 & 68.75 \\

                & Stacked cup separation
                & 27.50 & 32.50 & 41.25
                & 32.50 & 32.50 & 72.50
                & 58.75 & 85.00 & 86.25
                & 41.25 & 46.25 & 55.00 \\

                & Deformable cup transport
                & 11.00 & 23.00 & 26.00
                & 52.00 & 48.00 & 54.00
                & 37.00 & 58.00 & 90.00
                & 44.00 & 53.00 & 63.00 \\

                & Stiffness sorting
                & 30.00 & 22.50 & 47.50
                & 32.50 & 62.50 & 82.50
                & 67.50 & 50.00 & 65.00
                & 50.00 & 67.50 & 72.50 \\

                & Box handover
                & 4.00 & 6.00 & 28.00
                & 60.00 & 30.00 & 61.00
                & 50.00 & 54.00 & 74.00
                & 62.00 & 61.00 & 61.00 \\

                \cmidrule(l){2-14}
                & Mean success rate (\%)
                & 19.38 & 24.00 & \textbf{35.67}
                & 49.71 & 48.00 & \textbf{74.17}
                & 57.21 & 57.21 & \textbf{75.04}
                & 49.96 & 51.92 & \textbf{63.79} \\
                \bottomrule
            \end{tabular}%
\end{table*}

\begin{figure*}[!t]
    \centering
    \includegraphics[width=1.0\linewidth]{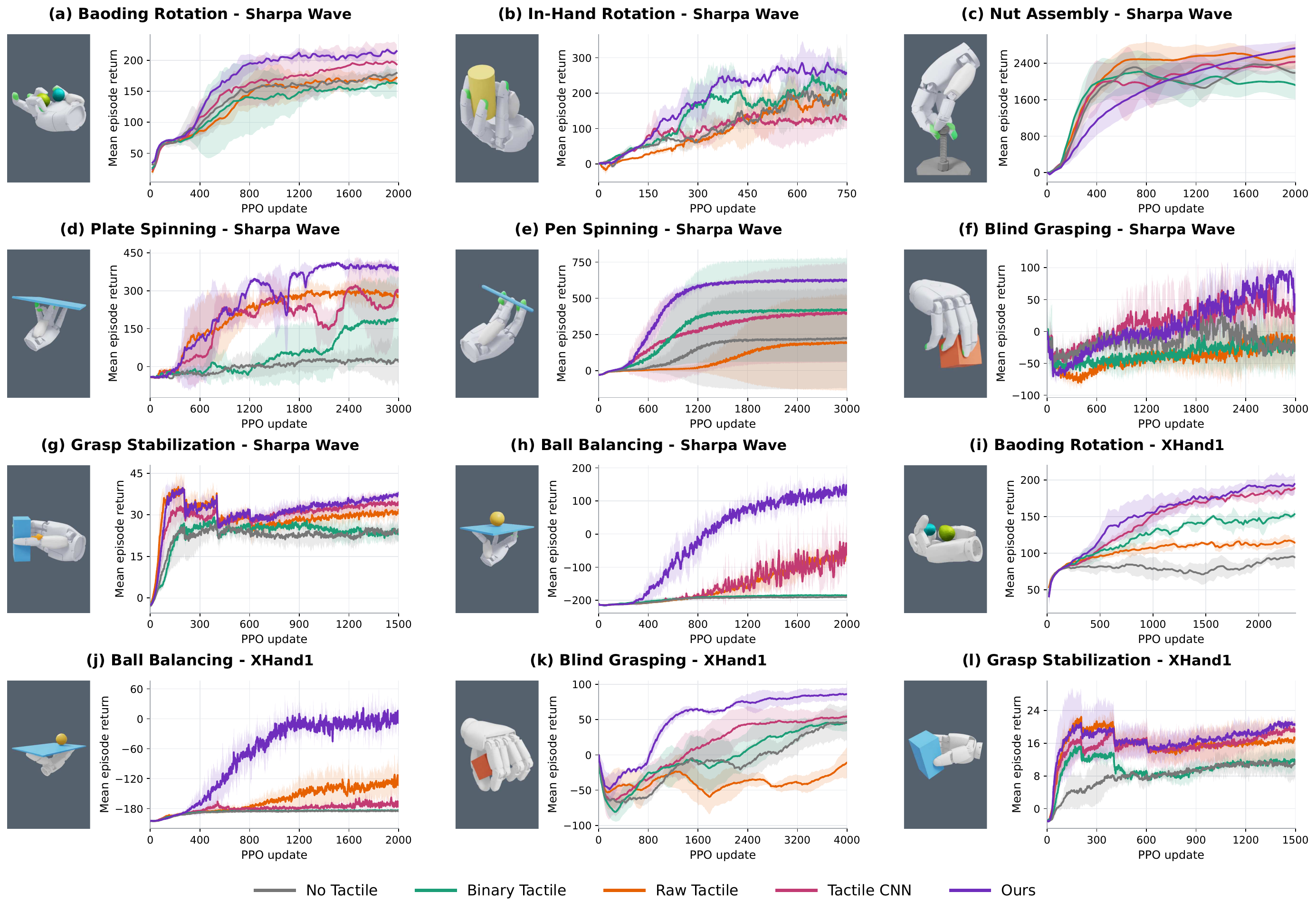}
    \caption{\textbf{Reinforcement-learning curves on twelve dexterous tasks}, Sharpa Wave (a--h) and XHand1 (i--l). Lines show the mean episode return over three seeds; bands show one standard deviation across the smoothed per-seed curves.}
    \label{rl}
\end{figure*}

\subsection{Reinforcement Learning in Simulation}
\label{sec:exp:robot:sim}

\paragraph{Final performance.}
\method features give the highest success rate and the highest return on all twelve tasks (\cref{tab:sim_rl_robot_tasks}). Averaged equally over tasks, success reaches 69.0\%, against 45.6\% for CNN features trained with the policy, 38.5\% for the raw signal, 30.3\% for binary contact, and 21.4\% with no touch. The ordering is itself informative: a learned representation beats the raw signal, and a pretrained one beats a representation learned inside the task. The margin is widest where contact changes continuously and the policy must keep up. Plate spinning reaches 97.4\%  compared with 82.9\% for the best alternative, while pen spinning reaches 97.1\% compared with 64.2\%. For ball balancing, the success rate reaches 67.7\% on Sharpa Wave versus 32.4\% for the best alternative, and 35.9\% on XHand1 versus 10.0\%. Binary contact and no touch score zero on both hands.

\paragraph{Stability.}
The standard deviation across seeds tells the other half of the story. For \method it never exceeds 9.2 points and averages 3.9; for the alternatives it reaches 55 on pen spinning and 40 on plate spinning, which means some seeds solve the task and others never do. The learning curves in~\cref{rl} show the same effect over time: with \method the return rises earlier on pen spinning, on ball balancing on both hands, and on XHand1 blind grasping, and the shaded bands stay narrow. Nut assembly is the exception, with a slower start before the highest final return. A frozen representation that already separates contact states lowers the burden on exploration, and policies that start from it converge both faster and more reliably.

\begin{table*}[!t]
    \centering
    \caption{\textbf{Reinforcement learning on contact-rich dexterous manipulation tasks in simulation.} Integrating \method substantially improves training success rates and total rewards, consistently outperforming the compared tactile and non-tactile baselines across tasks on Sharpa Wave and XHand1.}
    \label{tab:sim_rl_robot_tasks}
    \footnotesize
    \setlength{\tabcolsep}{3pt}
    \renewcommand{\arraystretch}{1.12}
    \begin{tabular}{@{}lllrrrrr@{}}
        \toprule
        Hand model & Task & Measure & No Tactile & Binary Tactile & Raw Tactile & Tactile CNN & \multicolumn{1}{c}{Ours} \\
        \midrule
        \multirow{16}{*}{Sharpa Wave} & \multirow{2}{*}{Baoding Rotation} & Success (\%) $\uparrow$ & $34.9 \pm 6.8$ & $38.2 \pm 11.4$ & $35.5 \pm 8.7$ & $44.8 \pm 9.3$ & $\mathbf{55.5} \pm \mathbf{1.8}$ \\
         & & Reward $\uparrow$ & $175.8 \pm 5.4$ & $162.9 \pm 22.9$ & $169.3 \pm 8.2$ & $195.7 \pm 32.6$ & $\mathbf{212.7} \pm \mathbf{0.3}$ \\
        \cmidrule(lr){2-8}
         & \multirow{2}{*}{In-Hand Rotation} & Success (\%) $\uparrow$ & $11.5 \pm 13.5$ & $18.3 \pm 31.7$ & $33.4 \pm 19.5$ & $26.6 \pm 23.1$ & $\mathbf{56.4} \pm \mathbf{7.5}$ \\
         & & Reward $\uparrow$ & $187.7 \pm 75.9$ & $208.3 \pm 75.7$ & $188.1 \pm 17.0$ & $127.9 \pm 36.8$ & $\mathbf{265.1} \pm \mathbf{28.0}$ \\
        \cmidrule(lr){2-8}
         & \multirow{2}{*}{Nut Assembly} & Success (\%) $\uparrow$ & $28.3 \pm 19.9$ & $21.9 \pm 13.5$ & $55.8 \pm 7.5$ & $40.8 \pm 17.3$ & $\mathbf{86.5} \pm \mathbf{4.2}$ \\
         & & Reward $\uparrow$ & $2269.6 \pm 88.0$ & $1972.4 \pm 300.5$ & $2504.0 \pm 234.9$ & $2383.0 \pm 140.8$ & $\mathbf{2680.4} \pm \mathbf{176.1}$ \\
        \cmidrule(lr){2-8}
         & \multirow{2}{*}{Plate Spinning} & Success (\%) $\uparrow$ & $0.5 \pm 0.8$ & $56.5 \pm 40.5$ & $82.9 \pm 14.7$ & $59.9 \pm 36.7$ & $\mathbf{97.4} \pm \mathbf{2.0}$ \\
         & & Reward $\uparrow$ & $21.2 \pm 78.1$ & $180.3 \pm 218.3$ & $286.7 \pm 46.6$ & $248.8 \pm 151.3$ & $\mathbf{394.4} \pm \mathbf{17.7}$ \\
        \cmidrule(lr){2-8}
         & \multirow{2}{*}{Pen Spinning} & Success (\%) $\uparrow$ & $31.8 \pm 55.1$ & $64.2 \pm 55.6$ & $30.5 \pm 52.8$ & $62.1 \pm 54.0$ & $\mathbf{97.1} \pm \mathbf{1.1}$ \\
         & & Reward $\uparrow$ & $220.5 \pm 342.8$ & $417.7 \pm 356.2$ & $191.3 \pm 326.0$ & $396.1 \pm 339.5$ & $\mathbf{623.3} \pm \mathbf{3.0}$ \\
        \cmidrule(lr){2-8}
         & \multirow{2}{*}{Blind Grasping} & Success (\%) $\uparrow$ & $33.7 \pm 11.1$ & $30.8 \pm 18.3$ & $38.6 \pm 18.3$ & $67.6 \pm 12.4$ & $\mathbf{85.2} \pm \mathbf{9.2}$ \\
         & & Reward $\uparrow$ & $-22.6 \pm 17.9$ & $-23.2 \pm 37.8$ & $-15.0 \pm 41.6$ & $34.2 \pm 29.5$ & $\mathbf{78.5} \pm \mathbf{18.3}$ \\
        \cmidrule(lr){2-8}
         & \multirow{2}{*}{Grasp Stabilization} & Success (\%) $\uparrow$ & $35.2 \pm 14.4$ & $27.8 \pm 9.8$ & $63.3 \pm 9.0$ & $74.3 \pm 15.9$ & $\mathbf{80.3} \pm \mathbf{4.5}$ \\
         & & Reward $\uparrow$ & $23.9 \pm 3.3$ & $23.3 \pm 2.3$ & $30.6 \pm 2.2$ & $34.0 \pm 3.2$ & $\mathbf{36.6} \pm \mathbf{0.6}$ \\
        \cmidrule(lr){2-8}
         & \multirow{2}{*}{Ball Balancing} & Success (\%) $\uparrow$ & $0.0 \pm 0.0$ & $0.0 \pm 0.0$ & $32.4 \pm 4.0$ & $31.1 \pm 11.3$ & $\mathbf{67.7} \pm \mathbf{1.8}$ \\
         & & Reward $\uparrow$ & $-190.4 \pm 2.3$ & $-185.1 \pm 0.4$ & $-71.0 \pm 16.2$ & $-71.6 \pm 51.0$ & $\mathbf{127.2} \pm \mathbf{21.1}$ \\
        \midrule
        \multirow{8}{*}{XHand1} & \multirow{2}{*}{Baoding Rotation} & Success (\%) $\uparrow$ & $0.1 \pm 0.1$ & $24.2 \pm 2.6$ & $2.5 \pm 1.8$ & $30.4 \pm 6.6$ & $\mathbf{41.0} \pm \mathbf{2.5}$ \\
         & & Reward $\uparrow$ & $93.6 \pm 13.6$ & $151.0 \pm 5.7$ & $115.2 \pm 7.0$ & $185.2 \pm 6.4$ & $\mathbf{192.9} \pm \mathbf{14.2}$ \\
        \cmidrule(lr){2-8}
         & \multirow{2}{*}{Ball Balancing} & Success (\%) $\uparrow$ & $0.0 \pm 0.0$ & $0.0 \pm 0.0$ & $10.0 \pm 5.9$ & $0.7 \pm 0.6$ & $\mathbf{35.9} \pm \mathbf{4.9}$ \\
         & & Reward $\uparrow$ & $-184.9 \pm 0.3$ & $-183.7 \pm 0.4$ & $-128.8 \pm 27.9$ & $-171.9 \pm 6.1$ & $\mathbf{-2.8} \pm \mathbf{17.0}$ \\
        \cmidrule(lr){2-8}
         & \multirow{2}{*}{Blind Grasping} & Success (\%) $\uparrow$ & $75.0 \pm 8.7$ & $72.4 \pm 4.0$ & $55.1 \pm 7.2$ & $77.0 \pm 4.9$ & $\mathbf{87.8} \pm \mathbf{4.2}$ \\
         & & Reward $\uparrow$ & $46.0 \pm 25.0$ & $45.1 \pm 10.4$ & $-17.7 \pm 15.9$ & $53.1 \pm 10.6$ & $\mathbf{85.5} \pm \mathbf{8.5}$ \\
        \cmidrule(lr){2-8}
         & \multirow{2}{*}{Grasp Stabilization} & Success (\%) $\uparrow$ & $6.0 \pm 3.2$ & $9.7 \pm 5.9$ & $21.7 \pm 12.7$ & $31.8 \pm 2.1$ & $\mathbf{36.6} \pm \mathbf{3.4}$ \\
         & & Reward $\uparrow$ & $10.9 \pm 3.1$ & $11.5 \pm 2.3$ & $16.6 \pm 4.8$ & $18.9 \pm 0.7$ & $\mathbf{20.5} \pm \mathbf{0.7}$ \\
        \bottomrule
    \end{tabular}%
\end{table*}

\subsection{Which Representation Helps a Policy}
\label{sec:exp:robot:which}
\begin{figure*}[!t]
    \centering
    \begin{subfigure}[b]{0.40\linewidth}
        \centering
        \includegraphics[width=\linewidth]{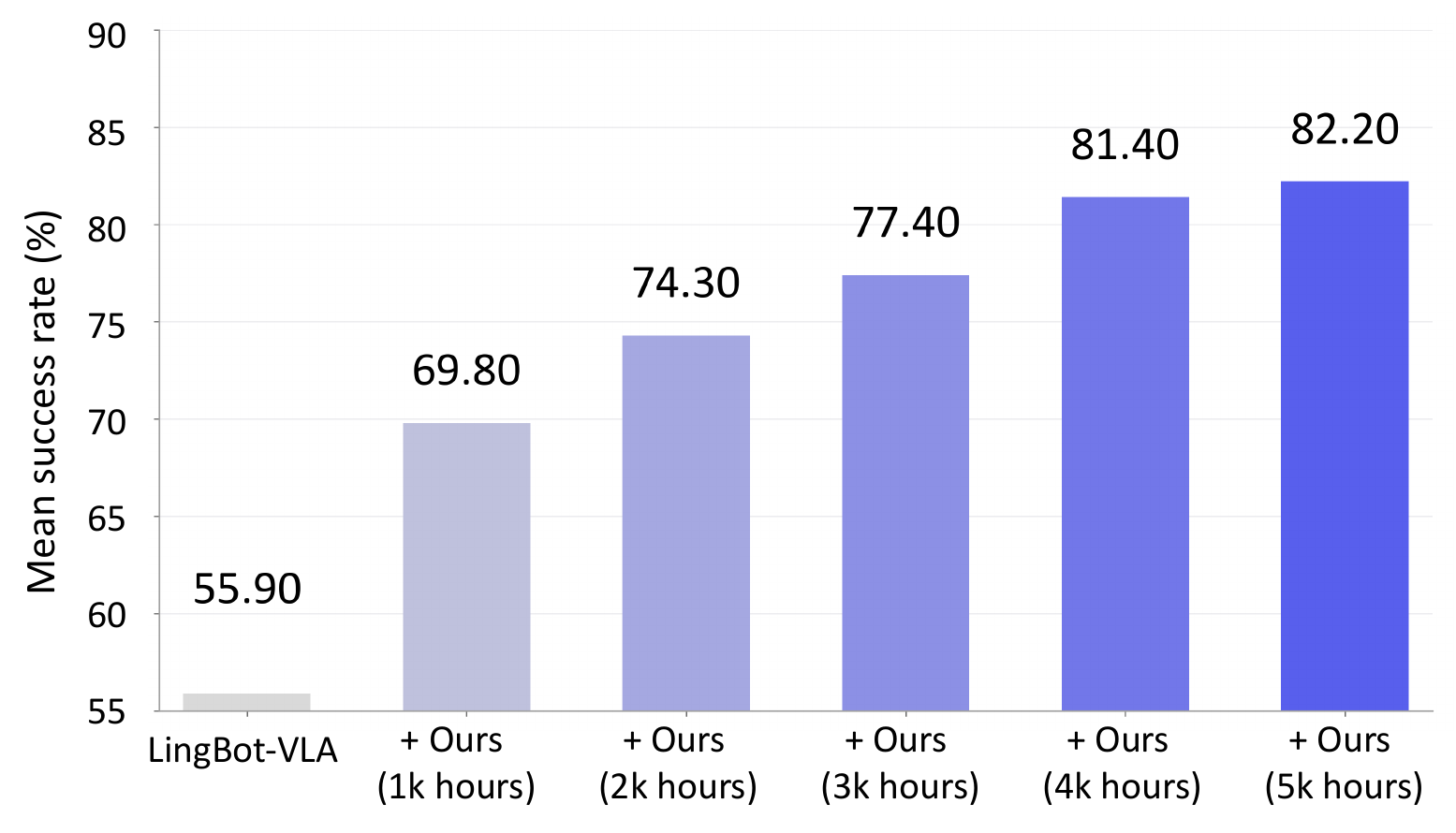}
        \caption{Pretraining hours.}
        \label{fig:which:scaling}
    \end{subfigure}
    \hfill
    \begin{subfigure}[b]{0.58\linewidth}
        \centering
        \includegraphics[width=\linewidth]{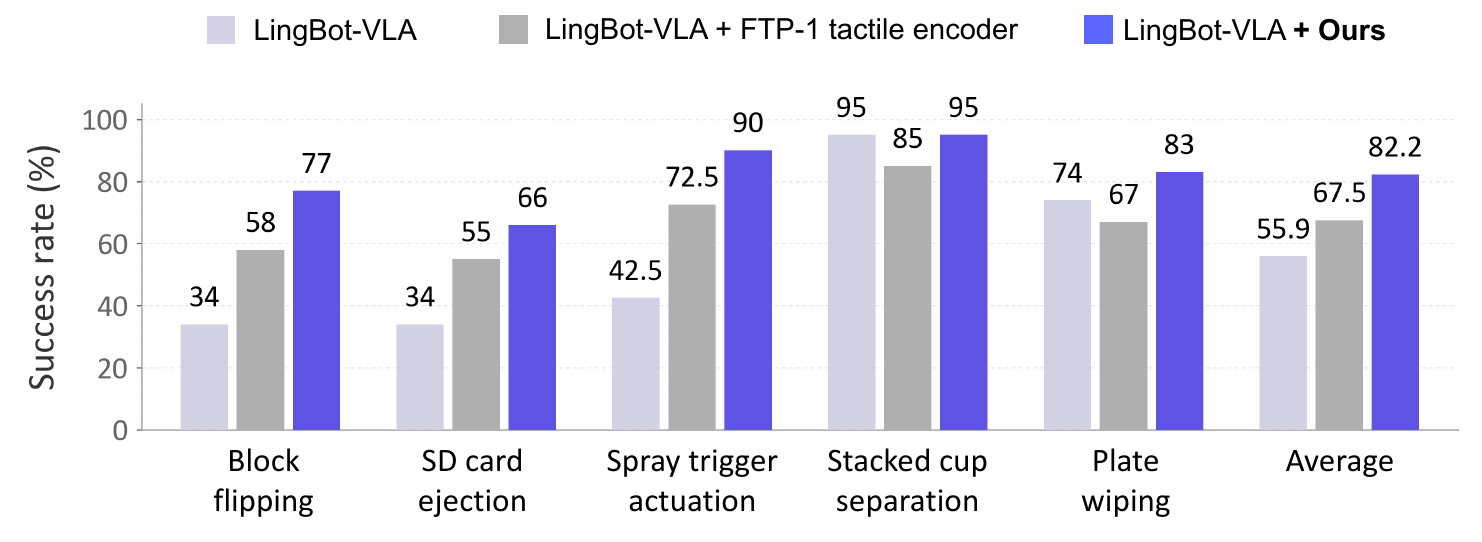}
        \caption{Tactile encoder.}
        \label{fig:which:ftp1}
    \end{subfigure}
    \caption{\textbf{Which representation helps a policy?} We evaluate \lingbotvla{} post-trained on five Sharpa Wave tasks, with 20 trials per task. (a) Mean success rate as the \method{} encoder's pretraining data increases from 1K to 5K hours. (b) Per-task success rates with the FTP-1 tactile encoder versus \method{} in the same policy. \method{} uses the HaT-Tiny configuration and is pretrained on the full 5K-hour dataset.}
    \label{fig:which}
\end{figure*}

The probing results of~\cref{sec:exp:probe} rank representations; this subsection examines whether these rankings translate into policy performance. Using the same \lingbotvla{} policy backbone, we post-train each configuration with 200 episodes per task and evaluate it over 20 trials on each of five Sharpa Wave tasks: block flipping, SD card ejection, spray trigger actuation, stacked cup separation, and plate wiping. The policy alone achieves an average success rate of 55.9\%, which increases to 82.2\% when augmented with tactile representations from the HaT-Tiny of \method (\cref{fig:which}).

\paragraph{Pretraining hours.}
Encoders pretrained on 1K to 5K hours lift the policy to 69.8\%, 74.3\%, 77.4\%, 81.4\%, and 82.2\% (\cref{fig:which}, left). Every additional thousand hours of tactile data, none of it from these tasks or this robot's demonstrations, makes the downstream policy better, and the curve has the same shape as the probing curves of~\cref{sec:exp:scalingtransfer}.

\paragraph{Another encoder.}
Replacing \method with the tactile encoder of FTP-1 in the same policy yields 67.5\%, less than half the gain (\cref{fig:which}, right). FTP-1 helps on three tasks but lowers success on stacked cup separation, 95\% to 85\%, and plate wiping, 74\% to 67\%; \method is at or above the original policy on all five. The probing gaps of~\cref{sec:ContactPhysics}, where FTP-1's errors are several times larger, show up as policy gaps.

\section{Conclusion and Discussion}
\label{sec:conclusion}
We presented \method, a tactile foundation model that learns a shared representation of contact from piezoresistive, magnetic, and visuotactile sensors. A canonical fingertip space gives heterogeneous sensors one input interface, and \obj trains a shared Haptic Transformer by predicting contact in latent space. Pretrained once on 5{,}000 hours of recordings from tactile gloves and teleoperated robot hands, the encoder is frozen and used as is.

Three findings follow. The representation belongs to contact rather than to a transducer: without any parameter update, the encoder recovers contact physics on a magnetic hand absent from pretraining and outperforms FTP-1 fitted with a sensor-specific adapter. The gain for robot learning comes from the representation, not the modality: frozen tokens improve imitation learning, VLA, and reinforcement learning policies across 26 real-world and twelve simulated tasks, while the raw signal helps at most half as much and often hurts. And the representation keeps improving with hours, sensor families, and model size, where signal reconstruction stalls.

These results suggest how to scale touch. No single sensor has thousands of hours, and it does not need them: intensity and dynamics transfer across sensing principles, so data can be collected with whatever sensor is easiest to record and spent on whatever hand is deployed, at some cost in spatial precision. We hope this informs the design of future tactile hardware and of embodied models that treat touch as a first-class input.

\paragraph{Limitations.}
First, real-world and simulation experiments need to be conducted using a wider variety of hand embodiments. Second, sensor coverage can be extended further. Capacitive sensors are currently absent from pretraining because the capacitive recordings available to us did not meet our quality requirements. Finally, the canonical geometry and pretraining layouts focus on the fingertips, leaving contacts on the palm and elsewhere along the fingers underrepresented. Extending layout coverage from the fingertips to the whole hand is an important next step.

\justifying

\FloatBarrier

\newpage

\section*{Appendix}
\appendix
\section{Implementation Details}
\label{app:impl}

This section provides additional details on the pretraining corpus and the frozen-encoder probing setup. \cref{tab:corpus} summarizes the data sources, including their sensing principles, collection platforms and modes, dataset sizes, spatial resolutions, and sampling rates. \cref{tab:probe_tasks} specifies the five contact-physics probing tasks, together with their input and output representations, training and test data durations, and evaluation metrics. \cref{tab:probe_hyperparameters} reports the shared probe architecture and optimization settings, as well as task-specific losses, batch sizes, and additional configurations. Throughout probing, the pretrained target encoder remains frozen, and probes are trained separately for Sharpa Wave and XHand1.

\begin{table}[h]
  \centering
  \caption{\textbf{Composition of the pretraining corpus.} Each source is listed with its sensing principle, platform, collection mode, and scale. $^{\dagger}$TACHIN uses 124 valid taxels on the thumb and 61 on each remaining finger. PaXini uses 116 valid taxels on the thumb and 96 on each remaining finger.}
  \label{tab:corpus}
  \footnotesize
  \setlength{\tabcolsep}{4pt}
  \renewcommand{\arraystretch}{1}

  \begin{tabular}{@{}llllrrrr@{}}
    \toprule
    Sensing & Sensor / vendor & Platform & Collection
    & Hours & Episodes & Taxels/finger & Rate (Hz) \\
    \midrule

    \multirow{3}{*}{Piezoresistive}
      & Juqiao
      & \multirow{3}{*}{\shortstack[l]{Tactile glove\\(human)}}
      & \multirow{3}{*}{Egocentric}
      & \multirow{3}{*}{2{,}197}
      & \multirow{3}{*}{86{,}906}
      & 9 & 100 \\
      & Xspark AI & & & & & 16 & 60 \\
      & TACHIN & & & & & 124 / 61$^{\dagger}$ & 60 \\
    \midrule

    Magnetic
      & PaXini
      & \shortstack[l]{Tactile glove\\(human)}
      & Egocentric
      & 1{,}136
      & 294{,}871
      & 116/96 $^{\dagger}$
      & 41.7-83.3 \\
    \midrule

    Visuotactile
      & \shortstack[l]{Sharpa Wave\\fingertip}
      & Dual-arm robot
      & Teleoperation
      & 1{,}650
      & 82{,}654
      & 240×240
      & 30 \\
    \midrule

    Total & & & & 4{,}983 & 464{,}431 & & \\
    \bottomrule
  \end{tabular}
\end{table}

\begin{table}[h]
  \centering
  \caption{\textbf{Contact-physics probing tasks.} Training and test
  durations, input and output specifications, and evaluation metrics
  for frozen-encoder probing.}
  \label{tab:probe_tasks}
  \footnotesize
  \setlength{\tabcolsep}{3pt}
  \renewcommand{\arraystretch}{1.2}
  \begin{tabular}{@{}lllrrl@{}}
    \toprule
    Task & Input & Output & Train (h) & Test (h) & Metric \\
    \midrule
    Fingertip force recovery
      & \shortstack[l]{Tactile clip\\$16\times5\times M\times3$}
      & \shortstack[l]{Force magnitudes\\$8\times5\times1$}
      & 2.492 & 0.361
      & Force magnitude MAE (N) \\
    Contact center estimation
      & \shortstack[l]{Tactile clip\\$16\times5\times M\times3$}
      & \shortstack[l]{3D contact centers\\$8\times5\times3$}
      & 2.616 & 0.334
      & Distance error (mm) \\
    Contact area estimation
      & \shortstack[l]{Tactile clip\\$16\times5\times M\times3$}
      & \shortstack[l]{Coverage ratios\\$8\times5\times1$}
      & 2.616 & 0.334
      & Coverage MAE (\%) \\
    Future force prediction
      & \shortstack[l]{Tactile clip + joint positions\\
          $16\times5\times M\times3;\;5\times J$}
      & \shortstack[l]{Future force magnitudes\\$4\times5\times1$}
      & 1.843 & 0.484
      & Force magnitude MAE (N) \\
    Tactile-to-video retrieval
      & \shortstack[l]{Tactile clip\\$16\times5\times M\times3$}
      & Similarity scores
      & 1.798 & 0.237
      & Top-5 recall (\%) \\
    \bottomrule
  \end{tabular}
\end{table}

\begin{table*}[!t]
  \centering
  \caption{Shared training hyperparameters and task-specific settings for frozen-encoder probing. Sharpa Wave and XHand1 probes are trained separately.}
  \label{tab:probe_hyperparameters}
  \begingroup
  \footnotesize
  \setlength{\tabcolsep}{4pt}
  \renewcommand{\arraystretch}{1.15}
  \begin{tabularx}{\textwidth}{@{}>{\raggedright\arraybackslash}p{0.37\textwidth}>{\raggedright\arraybackslash}X@{}}
    \toprule
    \multicolumn{2}{@{}l}{\textbf{Shared settings}} \\
    \midrule
    Tactile encoder & Frozen pretrained target encoder \\
    Tactile input & 16 frames; 5 fingers of one hand \\
    Readout blocks / width / heads & 1 / 192 / 3 \\
    MLP hidden width / dropout & 768 / 0 \\
    Optimizer / learning rate / weight decay & AdamW / $10^{-4}$ / $0.01$ \\
    Training epochs / GPUs per run & 100 / 8 \\
    Learning-rate schedule / warm-up & Constant / none \\
    \midrule
  \end{tabularx}
  \par\vspace{2pt}
  \begin{tabularx}{\textwidth}{@{}>{\raggedright\arraybackslash}p{0.16\textwidth}>{\raggedright\arraybackslash}p{0.15\textwidth}>{\centering\arraybackslash}p{0.07\textwidth}>{\raggedright\arraybackslash}p{0.18\textwidth}>{\raggedright\arraybackslash}X@{}}
    \multicolumn{5}{@{}l}{\textbf{Task-specific settings}} \\
    \midrule
    Task & Loss & Global batch & Metric & Additional settings \\
    \midrule
    Force recovery & Magnitude L1 & 128 & Magnitude MAE (N) & Attentive readout. \\
    \addlinespace[4pt]
    Contact center & Euclidean distance & 128 & Mean distance (mm) & Per-finger training-set z-score coordinates. \\
    \addlinespace[4pt]
    Contact coverage & Coverage L1 & 128 & Coverage MAE (pp) & Sigmoid output; targets in $[0,1]$. \\
    \addlinespace[4pt]
    Future force prediction & Magnitude L1 & 128 & 4-frame magnitude MAE (N) & Measured current/future joint positions and last force magnitude; state/fusion widths 64/64. \\
    \addlinespace[4pt]
    Tactile--video retrieval & Symmetric InfoNCE & 64 pairs & Recall@5 (\%) & Frozen V-JEPA 2 ViT-L; 16 video frames at $256\times256$; projection dimension 1024; $\tau=0.07$. \\
    \bottomrule
  \end{tabularx}
  \par\vspace{4pt}
  \begin{minipage}{\textwidth}
    \footnotesize\raggedright
    \textit{Note:} Center and coverage share one probe with equal loss weights. Retrieval batches use one pair per recording and share negatives across GPUs. pp: percentage points.
  \end{minipage}
  \endgroup
\end{table*}

\FloatBarrier

\section{Real-Robot Evaluation}
\label{app:real_robot_tasks}

\begin{figure*}[htbp]
    \centering
    \includegraphics[width=\textwidth]{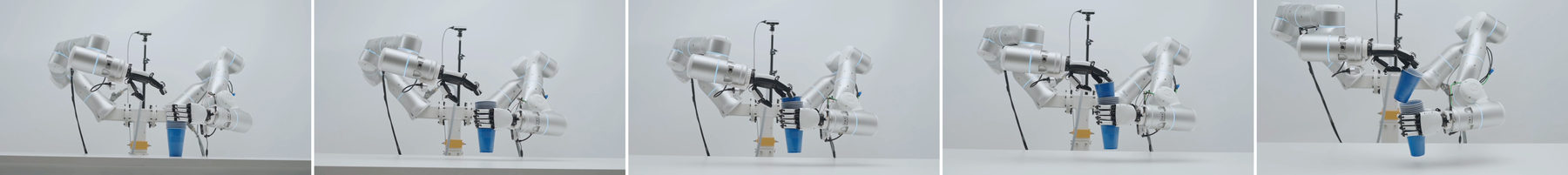}
    \caption{\textbf{Stacked cup separation (XHand1).} The robot stabilizes the nested stack, grasps and separates the uppermost cup, and holds the separated cup independently.}
    \label{fig:real_robot_xhand_stacked_cup_separation}
\end{figure*}

\begin{figure*}[htbp]
    \centering
    \includegraphics[width=\textwidth]{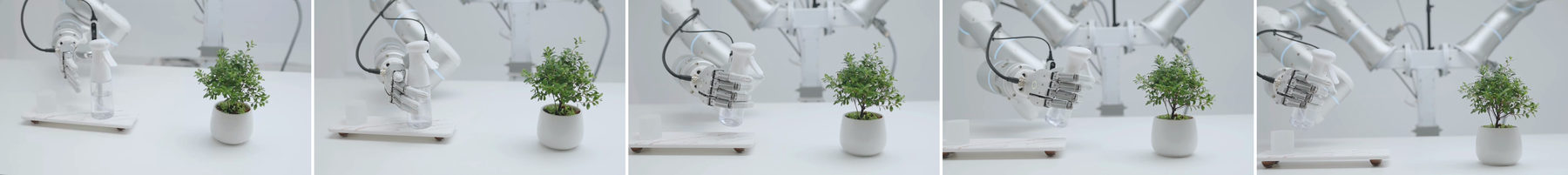}
    \caption{\textbf{Spray trigger actuation (Sharpa Wave).} The robot grasps the spray bottle, aims the nozzle toward the target, and actuates the trigger to discharge water.}
    \label{fig:real_robot_sharpa_wave_spray_trigger_actuation}
\end{figure*}

\cref{tab:real_robot_scoring} summarizes task descriptions, hand coverage, and scoring criteria. Check marks indicate evaluated hand configurations. Numbers in parentheses denote points; allocations are additive unless an ordinal scale is stated, in which case the highest level reached determines the score.~\cref{fig:real_robot_xhand_stacked_cup_separation,fig:real_robot_sharpa_wave_spray_trigger_actuation} illustrate two of the evaluated tasks. Each image contains five representative frames ordered from left to right.

\begin{table*}[!b]
    \centering
    \caption{Task descriptions, hand coverage, and scoring criteria for the real-robot evaluation.}
    \label{tab:real_robot_scoring}
    \begingroup
    \footnotesize
    \setlength{\tabcolsep}{1.5pt}
    \renewcommand{\arraystretch}{1.12}
    \begin{tabular}{@{}lccll@{}}
    \noalign{\hrule height 0.7pt}
    \parbox[c]{0.14\textwidth}{Task} &
    \makebox[0.09\textwidth][c]{Sharpa Wave} &
    \makebox[0.06\textwidth][c]{XHand1} &
    \parbox[c]{0.19\textwidth}{Description} &
    \parbox[c]{\dimexpr0.52\textwidth-8\tabcolsep\relax}{Success criterion} \\
    \noalign{\hrule height 0.25pt}
    \parbox[c]{0.14\textwidth}{\raggedright\vspace{1.8pt}\strut Single-page turning\strut\par\vspace{1.8pt}} &
    \makebox[0.09\textwidth][c]{$\checkmark$} &
    \makebox[0.06\textwidth][c]{$\checkmark$} &
    \parbox[c]{0.19\textwidth}{\raggedright\vspace{1.8pt}\strut Turn one page right to left.\strut\par\vspace{1.8pt}} &
    \parbox[c]{\dimexpr0.52\textwidth-8\tabcolsep\relax}{\raggedright\vspace{1.8pt}\strut 1 point each: approach edge, move page by contact, cross spine with one page, complete turn. No tearing. Total 4.\strut\par\vspace{1.8pt}} \\[0pt]
    \noalign{\hrule height 0.25pt}
    \parbox[c]{0.14\textwidth}{\raggedright\vspace{1.8pt}\strut Plate wiping\strut\par\vspace{1.8pt}} &
    \makebox[0.09\textwidth][c]{$\checkmark$} &
    \makebox[0.06\textwidth][c]{$\mbox{---}$} &
    \parbox[c]{0.19\textwidth}{\raggedright\vspace{1.8pt}\strut Wipe a stain from a stabilized plate.\strut\par\vspace{1.8pt}} &
    \parbox[c]{\dimexpr0.52\textwidth-8\tabcolsep\relax}{\raggedright\vspace{1.8pt}\strut 1 point each: lift cloth; stabilize plate; touch stain; clean; return cloth and withdraw hands. Plate intact and stable. Total 5.\strut\par\vspace{1.8pt}} \\[0pt]
    \noalign{\hrule height 0.25pt}
    \parbox[c]{0.14\textwidth}{\raggedright\vspace{1.8pt}\strut Stacked cup separation\strut\par\vspace{1.8pt}} &
    \makebox[0.09\textwidth][c]{$\checkmark$} &
    \makebox[0.06\textwidth][c]{$\checkmark$} &
    \parbox[c]{0.19\textwidth}{\raggedright\vspace{1.8pt}\strut Separate the uppermost nested cup.\strut\par\vspace{1.8pt}} &
    \parbox[c]{\dimexpr0.52\textwidth-8\tabcolsep\relax}{\raggedright\vspace{1.8pt}\strut 1 point each: stabilize stack; grasp top cup; separate; hold 2~s. Exactly one cup, no damage or stack disturbance. Total 4.\strut\par\vspace{1.8pt}} \\[0pt]
    \noalign{\hrule height 0.25pt}
    \parbox[c]{0.14\textwidth}{\raggedright\vspace{1.8pt}\strut Mahjong sorting\strut\par\vspace{1.8pt}} &
    \makebox[0.09\textwidth][c]{$\checkmark$} &
    \makebox[0.06\textwidth][c]{$\mbox{---}$} &
    \parbox[c]{0.19\textwidth}{\raggedright\vspace{1.8pt}\strut Sort a mahjong tile by touch.\strut\par\vspace{1.8pt}} &
    \parbox[c]{\dimexpr0.52\textwidth-8\tabcolsep\relax}{\raggedright\vspace{1.8pt}\strut Grasp (1); correct tray (+3) or wrong tray (+1); maximum 4. Face exposure to the policy camera during identification invalidates the trial.\strut\par\vspace{1.8pt}} \\[0pt]
    \noalign{\hrule height 0.25pt}
    \parbox[c]{0.14\textwidth}{\raggedright\vspace{1.8pt}\strut Pillbox refilling\strut\par\vspace{1.8pt}} &
    \makebox[0.09\textwidth][c]{$\checkmark$} &
    \makebox[0.06\textwidth][c]{$\mbox{---}$} &
    \parbox[c]{0.19\textwidth}{\raggedright\vspace{1.8pt}\strut Refill a compartment and close its lid.\strut\par\vspace{1.8pt}} &
    \parbox[c]{\dimexpr0.52\textwidth-8\tabcolsep\relax}{\raggedright\vspace{1.8pt}\strut 1 point each: approach latch; release; grasp surrogate; deposit; close lid. Surrogate retained, lid latched. Total 5.\strut\par\vspace{1.8pt}} \\[0pt]
    \noalign{\hrule height 0.25pt}
    \parbox[c]{0.14\textwidth}{\raggedright\vspace{1.8pt}\strut Business card extraction\strut\par\vspace{1.8pt}} &
    \makebox[0.09\textwidth][c]{$\checkmark$} &
    \makebox[0.06\textwidth][c]{$\mbox{---}$} &
    \parbox[c]{0.19\textwidth}{\raggedright\vspace{1.8pt}\strut Extract and retain a business card.\strut\par\vspace{1.8pt}} &
    \parbox[c]{\dimexpr0.52\textwidth-8\tabcolsep\relax}{\raggedright\vspace{1.8pt}\strut 1 point each: align; pinch; extract; hold stably after extraction. Total 4.\strut\par\vspace{1.8pt}} \\[0pt]
    \noalign{\hrule height 0.25pt}
    \parbox[c]{0.14\textwidth}{\raggedright\vspace{1.8pt}\strut Card dealing\strut\par\vspace{1.8pt}} &
    \makebox[0.09\textwidth][c]{$\checkmark$} &
    \makebox[0.06\textwidth][c]{$\mbox{---}$} &
    \parbox[c]{0.19\textwidth}{\raggedright\vspace{1.8pt}\strut Separate and lift one playing card.\strut\par\vspace{1.8pt}} &
    \parbox[c]{\dimexpr0.52\textwidth-8\tabcolsep\relax}{\raggedright\vspace{1.8pt}\strut Ordinal: approach (1), contact (2), slide out (3), grasp and lift (4).\strut\par\vspace{1.8pt}} \\[0pt]
    \noalign{\hrule height 0.25pt}
    \parbox[c]{0.14\textwidth}{\raggedright\vspace{1.8pt}\strut Deformable cup transport\strut\par\vspace{1.8pt}} &
    \makebox[0.09\textwidth][c]{$\checkmark$} &
    \makebox[0.06\textwidth][c]{$\checkmark$} &
    \parbox[c]{0.19\textwidth}{\raggedright\vspace{1.8pt}\strut Transport a thin-walled cup without damage.\strut\par\vspace{1.8pt}} &
    \parbox[c]{\dimexpr0.52\textwidth-8\tabcolsep\relax}{\raggedright\vspace{1.8pt}\strut Credit is based on contact, grasp stability, transport, deformation control, placement. Full credit requires a grasp of the outer sidewalls, no excessive deformation, and upright release in the target plate.\strut\par\vspace{1.8pt}} \\[0pt]
    \noalign{\hrule height 0.25pt}
    \parbox[c]{0.14\textwidth}{\raggedright\vspace{1.8pt}\strut SD card ejection\strut\par\vspace{1.8pt}} &
    \makebox[0.09\textwidth][c]{$\checkmark$} &
    \makebox[0.06\textwidth][c]{$\mbox{---}$} &
    \parbox[c]{0.19\textwidth}{\raggedright\vspace{1.8pt}\strut Eject and remove an SD card shell.\strut\par\vspace{1.8pt}} &
    \parbox[c]{\dimexpr0.52\textwidth-8\tabcolsep\relax}{\raggedright\vspace{1.8pt}\strut Approach (1); press and withdraw finger to unlock (2); grasp and remove shell completely (2). Total 5.\strut\par\vspace{1.8pt}} \\[0pt]
    \noalign{\hrule height 0.25pt}
    \parbox[c]{0.14\textwidth}{\raggedright\vspace{1.8pt}\strut Block flipping\strut\par\vspace{1.8pt}} &
    \makebox[0.09\textwidth][c]{$\checkmark$} &
    \makebox[0.06\textwidth][c]{$\mbox{---}$} &
    \parbox[c]{0.19\textwidth}{\raggedright\vspace{1.8pt}\strut Reorient a cube to the target face.\strut\par\vspace{1.8pt}} &
    \parbox[c]{\dimexpr0.52\textwidth-8\tabcolsep\relax}{\raggedright\vspace{1.8pt}\strut Grasp (1); finger reorientation (2); container placement (1); target face up (1). Wrist rotation alone earns no reorientation credit. Total 5.\strut\par\vspace{1.8pt}} \\[0pt]
    \noalign{\hrule height 0.25pt}
    \parbox[c]{0.14\textwidth}{\raggedright\vspace{1.8pt}\strut Pen uncapping\strut\par\vspace{1.8pt}} &
    \makebox[0.09\textwidth][c]{$\checkmark$} &
    \makebox[0.06\textwidth][c]{$\mbox{---}$} &
    \parbox[c]{0.19\textwidth}{\raggedright\vspace{1.8pt}\strut Remove a pen cap bimanually.\strut\par\vspace{1.8pt}} &
    \parbox[c]{\dimexpr0.52\textwidth-8\tabcolsep\relax}{\raggedright\vspace{1.8pt}\strut 1 point each: hold barrel; lift between hands; approach cap; grasp cap stably; remove and retain cap. Control both parts. Total 5.\strut\par\vspace{1.8pt}} \\[0pt]
    \noalign{\hrule height 0.25pt}
    \parbox[c]{0.14\textwidth}{\raggedright\vspace{1.8pt}\strut Cap twisting\strut\par\vspace{1.8pt}} &
    \makebox[0.09\textwidth][c]{$\checkmark$} &
    \makebox[0.06\textwidth][c]{$\mbox{---}$} &
    \parbox[c]{0.19\textwidth}{\raggedright\vspace{1.8pt}\strut Unscrew and remove a bottle cap.\strut\par\vspace{1.8pt}} &
    \parbox[c]{\dimexpr0.52\textwidth-8\tabcolsep\relax}{\raggedright\vspace{1.8pt}\strut Ordinal: approach (1), grasp and rotate effectively (2), fully unscrew (3), lift clear (4).\strut\par\vspace{1.8pt}} \\[0pt]
    \noalign{\hrule height 0.25pt}
    \parbox[c]{0.14\textwidth}{\raggedright\vspace{1.8pt}\strut Bag extraction\strut\par\vspace{1.8pt}} &
    \makebox[0.09\textwidth][c]{$\checkmark$} &
    \makebox[0.06\textwidth][c]{$\mbox{---}$} &
    \parbox[c]{0.19\textwidth}{\raggedright\vspace{1.8pt}\strut Pull a bag from its dispenser.\strut\par\vspace{1.8pt}} &
    \parbox[c]{\dimexpr0.52\textwidth-8\tabcolsep\relax}{\raggedright\vspace{1.8pt}\strut 1 point each: stabilize box; approach bag; stable pinch; lift visibly out. Total 4.\strut\par\vspace{1.8pt}} \\[0pt]
    \noalign{\hrule height 0.25pt}
    \parbox[c]{0.14\textwidth}{\raggedright\vspace{1.8pt}\strut Straw insertion\strut\par\vspace{1.8pt}} &
    \makebox[0.09\textwidth][c]{$\checkmark$} &
    \makebox[0.06\textwidth][c]{$\mbox{---}$} &
    \parbox[c]{0.19\textwidth}{\raggedright\vspace{1.8pt}\strut Insert a straw through a sealed lid.\strut\par\vspace{1.8pt}} &
    \parbox[c]{\dimexpr0.52\textwidth-8\tabcolsep\relax}{\raggedright\vspace{1.8pt}\strut Credit is based on prescribed grasp, film contact and depression, puncture, prescribed depth. No buckling, tipping, or cup-bottom contact.\strut\par\vspace{1.8pt}} \\[0pt]
    \noalign{\hrule height 0.25pt}
    \parbox[c]{0.14\textwidth}{\raggedright\vspace{1.8pt}\strut Fan opening\strut\par\vspace{1.8pt}} &
    \makebox[0.09\textwidth][c]{$\checkmark$} &
    \makebox[0.06\textwidth][c]{$\mbox{---}$} &
    \parbox[c]{0.19\textwidth}{\raggedright\vspace{1.8pt}\strut Activate and orient a fan.\strut\par\vspace{1.8pt}} &
    \parbox[c]{\dimexpr0.52\textwidth-8\tabcolsep\relax}{\raggedright\vspace{1.8pt}\strut 1 point each: stabilize fan; touch button; press to activate; orient airflow to move flag. Total 4.\strut\par\vspace{1.8pt}} \\[0pt]
    \noalign{\hrule height 0.25pt}
    \parbox[c]{0.14\textwidth}{\raggedright\vspace{1.8pt}\strut Zipper closing\strut\par\vspace{1.8pt}} &
    \makebox[0.09\textwidth][c]{$\checkmark$} &
    \makebox[0.06\textwidth][c]{$\mbox{---}$} &
    \parbox[c]{0.19\textwidth}{\raggedright\vspace{1.8pt}\strut Close a pouch zipper.\strut\par\vspace{1.8pt}} &
    \parbox[c]{\dimexpr0.52\textwidth-8\tabcolsep\relax}{\raggedright\vspace{1.8pt}\strut 1 point each: lift and stabilize pouch; approach; grasp slider; close zipper; release into container. Zipper stays closed. Total 5.\strut\par\vspace{1.8pt}} \\[0pt]
    \noalign{\hrule height 0.25pt}
    \parbox[c]{0.14\textwidth}{\raggedright\vspace{1.8pt}\strut Spray trigger actuation\strut\par\vspace{1.8pt}} &
    \makebox[0.09\textwidth][c]{$\checkmark$} &
    \makebox[0.06\textwidth][c]{$\mbox{---}$} &
    \parbox[c]{0.19\textwidth}{\raggedright\vspace{1.8pt}\strut Aim and operate a spray bottle.\strut\par\vspace{1.8pt}} &
    \parbox[c]{\dimexpr0.52\textwidth-8\tabcolsep\relax}{\raggedright\vspace{1.8pt}\strut Grasp (1); aim (1); trigger stage: actuation (1) or visible water discharge (2). Maximum 4.\strut\par\vspace{1.8pt}} \\[0pt]
    \noalign{\hrule height 0.25pt}
    \parbox[c]{0.14\textwidth}{\raggedright\vspace{1.8pt}\strut Stiffness sorting\strut\par\vspace{1.8pt}} &
    \makebox[0.09\textwidth][c]{$\checkmark$} &
    \makebox[0.06\textwidth][c]{$\checkmark$} &
    \parbox[c]{0.19\textwidth}{\raggedright\vspace{1.8pt}\strut Probe and sort sponges by stiffness.\strut\par\vspace{1.8pt}} &
    \parbox[c]{\dimexpr0.52\textwidth-8\tabcolsep\relax}{\raggedright\vspace{1.8pt}\strut Credit is based on press--release, correct category, placement in the matching tray. Full credit requires agreement with the predefined sample category.\strut\par\vspace{1.8pt}} \\[0pt]
    \noalign{\hrule height 0.25pt}
    \parbox[c]{0.14\textwidth}{\raggedright\vspace{1.8pt}\strut Book extraction\strut\par\vspace{1.8pt}} &
    \makebox[0.09\textwidth][c]{$\checkmark$} &
    \makebox[0.06\textwidth][c]{$\mbox{---}$} &
    \parbox[c]{0.19\textwidth}{\raggedright\vspace{1.8pt}\strut Extract a tightly shelved book.\strut\par\vspace{1.8pt}} &
    \parbox[c]{\dimexpr0.52\textwidth-8\tabcolsep\relax}{\raggedright\vspace{1.8pt}\strut Edge contact (1); partial pull (2); thumb--finger grasp (1); fully lift clear (1). Total 5.\strut\par\vspace{1.8pt}} \\[0pt]
    \noalign{\hrule height 0.25pt}
    \parbox[c]{0.14\textwidth}{\raggedright\vspace{1.8pt}\strut Box handover\strut\par\vspace{1.8pt}} &
    \makebox[0.09\textwidth][c]{$\checkmark$} &
    \makebox[0.06\textwidth][c]{$\checkmark$} &
    \parbox[c]{0.19\textwidth}{\raggedright\vspace{1.8pt}\strut Transfer a box between hands.\strut\par\vspace{1.8pt}} &
    \parbox[c]{\dimexpr0.52\textwidth-8\tabcolsep\relax}{\raggedright\vspace{1.8pt}\strut Credit is based on independent lift, shared grasp, full support transfer, target placement. Full credit requires an independent receiving-hand hold and stable placement without dropping.\strut\par\vspace{1.8pt}} \\[0pt]
    \noalign{\hrule height 0.25pt}
    \parbox[c]{0.14\textwidth}{\raggedright\vspace{1.8pt}\strut Egg transfer\strut\par\vspace{1.8pt}} &
    \makebox[0.09\textwidth][c]{$\mbox{---}$} &
    \makebox[0.06\textwidth][c]{$\checkmark$} &
    \parbox[c]{0.19\textwidth}{\raggedright\vspace{1.8pt}\strut Transfer an egg replica between holders.\strut\par\vspace{1.8pt}} &
    \parbox[c]{\dimexpr0.52\textwidth-8\tabcolsep\relax}{\raggedright\vspace{1.8pt}\strut 1 point each: grasp; lift clear; position above target; release. Egg stably seated. Total 4.\strut\par\vspace{1.8pt}} \\[0pt]

    \noalign{\hrule height 0.7pt}
    \end{tabular}
    \endgroup

\end{table*}

\clearpage

\section{Simulation-Based Evaluation of Tactile Representations}
\label{app:simulation}

\subsection{Simulation Probing}
\label{app:sim_probing}

\begin{figure*}[!t]
    \centering
    \includegraphics[width=1.0\linewidth]{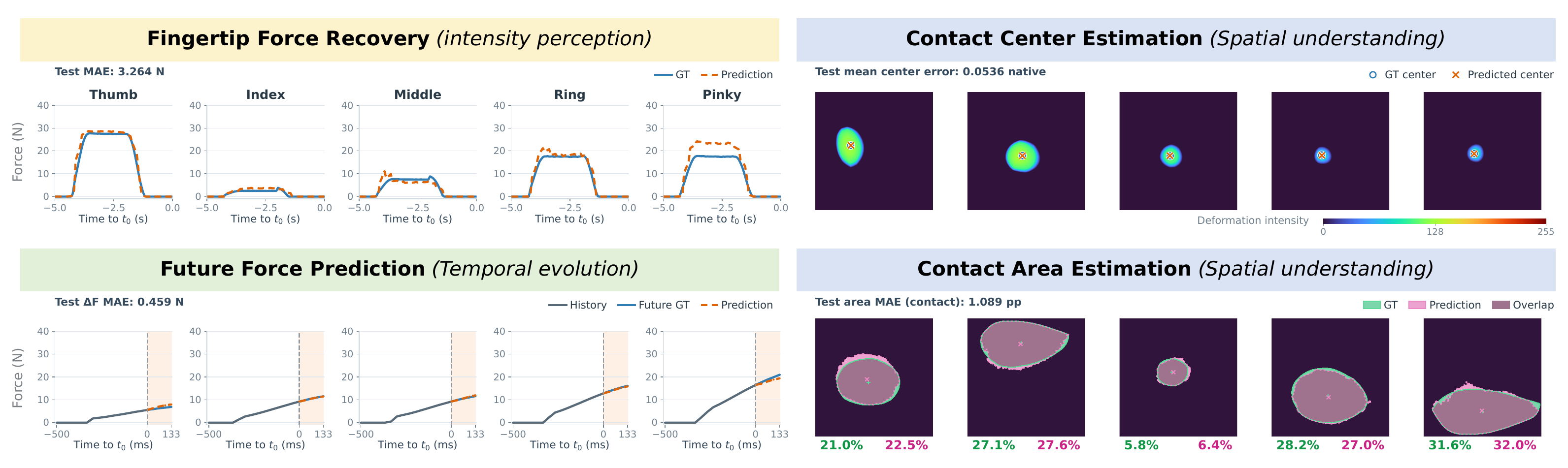}
    \caption{\textbf{Simulation-based contact physics probing.}
    We conduct the same probing tasks as in the real-world evaluation on the Sharpa Wave platform to assess the generalization of \methodname{} representations to simulated tactile data. The pretrained encoder remains frozen, and task-specific attentive probes are trained independently.}
    \label{sim_probe_setting}
\end{figure*}

Since \methodname{} is pretrained on multi-source real-world tactile data, its ability to generalize to simulated tactile observations remains to be evaluated. Before conducting reinforcement learning experiments, we perform simulation-based probing to assess whether the pretrained representations preserve contact-related physical information in the unseen simulation domain.
We collect 7{,}200 seconds of simulated single-hand grasping data on the Sharpa Wave platform, covering tactile interactions with 47 objects across diverse grasp poses. We conduct four probing tasks: fingertip force recovery, contact center estimation, contact area estimation, and future force prediction. We use the same training hyperparameters and evaluation metrics as in the real-world probing experiments. 

As shown in~\cref{sim_probe_setting}, the contact center and contact area prediction results indicate that the pretrained \method representations retain decodable spatial contact information in simulation. With the encoder frozen, the probe heads recover contact locations, spatial distributions, and coverage. Across the complete test set, contact-positive regions achieve a mean Dice score of 0.708 and an area MAE of 1.089 percentage points. These results suggest that \method representations support the recovery of simulated contact geometry despite the distribution shift between simulated tactile observations and real-world sensor measurements.

The fingertip force recovery and future force prediction probes further assess force-magnitude and temporal information. The current-force estimates capture loading, holding, and unloading trends, although deviations in absolute magnitude remain, with an overall MAE of 3.264 N. Future force-change prediction achieves an MAE of 0.459 N, indicating that historical tactile features contain information useful for predicting short-term force changes; the absolute future forces shown in the figure are reconstructed by adding the predicted changes to the observed current force. Differences in response mechanisms and contact dynamics between Isaac Sim's deformation-based tactile observations and real-world tactile measurements contribute to the absolute-force estimation errors in simulation. Overall, the four probing experiments indicate that frozen \method representations retain useful spatial contact, force-magnitude, and short-term temporal information in the simulation domain, providing a basis for using \method representations in subsequent simulation-based reinforcement learning experiments.

\subsection{Simulated Dexterous Manipulation Tasks}
\label{app:sim_tasks}

We construct eight types of dexterous manipulation tasks and conduct reinforcement learning experiments in simulation using two dexterous hands, Sharpa Wave and XHand1.~\cref{tab:sim_task_criteria} summarizes the hand-model coverage, description, and success criterion for each task.

\begin{table*}[!t]
\centering
\caption{Eight simulated task types, hand-model coverage, and episode-level success criteria.}
\label{tab:sim_task_criteria}
\begingroup
\footnotesize
\setlength{\tabcolsep}{4pt}
\setlength{\arrayrulewidth}{0.25pt}
\renewcommand{\arraystretch}{1.10}
\renewcommand{\tabularxcolumn}[1]{m{#1}}
\begin{tabularx}{\textwidth}{@{}lcc>{\raggedright\arraybackslash}X>{\raggedright\arraybackslash}X@{}}
\toprule[0.8pt]
Task & Sharpa Wave & XHand1 & Description & Success criterion \\
\midrule[0.5pt]
\mbox{Baoding Rotation} & \ding{51} & \ding{51} & Coordinate two objects of varying shapes and sizes within the palm to repeatedly exchange their target positions. & At least 10 target exchanges. \\
\cline{1-5}
\mbox{In-Hand Rotation} & \ding{51} & \textemdash & Coordinate finger motions to rotate objects spanning nine shapes and three scales about a prescribed axis. & At least 1 net turn about the prescribed axis. \\
\cline{1-5}
\mbox{Nut Assembly} & \ding{51} & \textemdash & Turn hexagonal nuts of four sizes through finger contacts under the thread constraint of a fixed bolt. & At least 5 net turns from the initial angle at episode end. \\
\cline{1-5}
\mbox{Plate Spinning} & \ding{51} & \textemdash & Support a square plate and spin it about the vertical axis while controlling tilt and maintaining supporting contacts. & Complete 3 valid net turns within 40~s. \\
\cline{1-5}
\mbox{Pen Spinning} & \ding{51} & \textemdash & Coordinate finger contacts to spin a slender capsule-shaped pen continuously while maintaining support and preventing drops. & Complete 3 valid net turns within 12~s. \\
\cline{1-5}
\mbox{Blind Grasping} & \ding{51} & \ding{51} & Without vision, locate, grasp, lift, and hold cubes and cylinders with varied sizes and randomized poses. & Maintain a lift of $\geq$6~cm with opposing contact for 2~s. \\
\cline{1-5}
\mbox{Grasp Stabilization} & \ding{51} & \ding{51} & Stabilize a grasped cuboid against external forces and torques while reducing contact load and preventing drops. & Retain for 12~s without failure; mean total normal load $\leq$10~N. \\
\cline{1-5}
\mbox{Ball Balancing} & \ding{51} & \ding{51} & Control tray orientation with the hand to balance a rolling ball and prevent it from falling. & Retain the ball for 10~s without task failure. \\
\bottomrule[0.8pt]
\end{tabularx}
\endgroup

\end{table*}

\subsection{Reinforcement Learning Experimental Settings}
\label{app:sim_rl_settings}

\noindent\textbf{Simulation-based Tactile Generation.}
We generate simulated tactile observations following the geometry-based penetration depth method of TacMap~\cite{tacmap2026}. Specifically, sensing points and their corresponding surface normals are predefined on the curved fingertip surfaces. During simulation, GPU-parallelized ray queries are performed against the object meshes to estimate local penetration depths along the surface normals. These measurements are then assembled into spatial deformation maps that represent the contact distribution and local fingertip deformation.

\noindent\textbf{Tactile Observation Processing.}
To accommodate different fingertip geometries, we construct tactile observations according to the sensing layouts of Sharpa Wave and XHand1. Specifically, Sharpa Wave uses 96 sensing points per finger, while XHand1 uses 120 sensing points per finger. The simulated deformation measurements are normalized, and 8 consecutive frames are stacked to capture temporal contact dynamics.

\noindent\textbf{Tactile Observation Configurations.}
To evaluate the contribution of different tactile representations to reinforcement learning, we consider five observation configurations: \textit{No Tactile}, \textit{Binary Tactile}, \textit{Raw Tactile History}, \textit{Tactile CNN}, and the pretrained \methodname{} representation. The No Tactile baseline uses only non-tactile observations, including proprioceptive information. Binary Tactile provides thresholded contact indicators derived from the deformation measurements, while Raw Tactile History directly uses the temporal sequence of deformation observations. Tactile CNN employs a trainable tactile encoder optimized jointly with the policy. For \methodname{}, the simulated deformation observations are processed by the frozen pretrained tactile encoder, followed by a trainable readout module that produces features for downstream control.
For the pretrained encoder, the scalar deformation measurements are placed in the third channel of the three-dimensional tactile input, while the first two channels are set to zero. The resulting tactile features are concatenated with the corresponding non-tactile observations and provided to the policy network.

\noindent\textbf{Reinforcement Learning Protocol.}
Within each manipulation task, all observation configurations are evaluated under the same reinforcement learning framework, with consistent action spaces, reward functions, environment randomization, and task-specific training settings. The primary difference between configurations lies in how tactile information is represented and processed before being incorporated into the policy. This controlled comparison allows us to assess the influence of tactile representations on policy learning efficiency, training stability, and final manipulation performance.

\FloatBarrier
{\small
\bibliographystyle{plain}
\bibliography{references}
}
\end{document}